\documentclass[letterpaper, 10 pt, conference]{ieeeconf}  

\usepackage{cite}
\usepackage{amsmath,amssymb,amsfonts}
\usepackage[pdftex]{graphicx}
\usepackage{textcomp}
\def\BibTeX{{\rm B\kern-.05em{\sc i\kern-.025em b}\kern-.08em
    T\kern-.1667em\lower.7ex\hbox{E}\kern-.125emX}}
    
\usepackage[utf8]{inputenc}

\usepackage{subfig}
\usepackage{bm}
\usepackage{comment}
\usepackage{nccmath}
\usepackage{amsfonts}
\usepackage{algorithm, algpseudocode}
\usepackage{cite}
\newcommand{\argmax}{\mathop{\mathrm{arg~max}}\limits}
\newcommand{\argmin}{\mathop{\mathrm{arg~min}}\limits}
\usepackage{url}

\title{\LARGE \bf
    Biosignal Generation and Latent Variable Analysis with Recurrent Generative Adversarial Networks}
    
\author{Shota Harada, Hideaki Hayashi, and Seiichi Uchida
}

\begin{document}
		\maketitle
	\thispagestyle{empty}
	\pagestyle{empty}

\begin{abstract}
    The effectiveness of biosignal generation and data augmentation with biosignal generative models based on generative adversarial networks (GANs), which are a type of deep learning technique, was demonstrated in our previous paper.
    GAN-based generative models only learn the projection between a random distribution as input data and the distribution of training data.
    Therefore, the relationship between input and generated data is unclear, and the characteristics of the data generated from this model cannot be controlled.
    This study proposes a method for generating time-series data based on GANs and explores their ability to generate biosignals with certain classes and characteristics.
    Moreover, in the proposed method, latent variables are analyzed using canonical correlation analysis (CCA) to represent the relationship between input and generated data as canonical loadings.
    Using these loadings, we can control the characteristics of the data generated by the proposed method.
    The influence of class labels on generated data is analyzed by feeding the data interpolated between two class labels into the generator of the proposed GANs.
    The CCA of the latent variables is shown to be an effective method of controlling the generated data characteristics. 
    We are able to model the distribution of the time-series data without requiring domain-dependent knowledge using the proposed method.
    Furthermore, it is possible to control the characteristics of these data by analyzing the model trained using the proposed method.
    To the best of our knowledge, this work is the first to generate biosignals using GANs while controlling the characteristics of the generated data.
    \end{abstract}

\section{Introduction}
Biosignals, such as electrocardiogram (ECG) and electroencephalogram (EEG) signals, strongly reflect human internal states.
In particular, abnormality in the human body, including diseases, can cause visible changes in the patterns of biosignals.
For example, myocardial infarction induces an increase in the Q-wave and ST segment of ECGs.
Therefore, abnormality in the human body can be detected by classifying the patterns of biosignals.
In fact, physicians refer to the patterns of biosignals to diagnose diseases and determine treatment.

Biosignal analysis benefits various fields such as medicine and healthcare.
In the medical field, biosignal analysis is utilized to detect diseases such as myocardial infarction \cite{kora2017}, epileptic seizures \cite{lina2017}, and psychiatric disorders \cite{Rostamabad2010}.
In healthcare applications, biosignal analysis is utilized for brain--computer interfaces (BCIs) and the control of prosthetic limbs based on electromyograms \cite{sudarsan2012}.
For BCIs, Rahul textit{et al.} have reported that electric wheelchairs are controlled using EEG \cite{yumlembam2017}.
In BCI applications in other fields, attempts to control drones using EEG have also been reported \cite{lafleur2013}.

Numerous studies have reported that biosignals can be identified using a discriminative model of deep learning \cite{Bahareh2017, Chambon2018}.
Owing to the development of deep learning, a few studies have achieved considerable increases in classification accuracy.

The study of generative models based on deep learning was motivated by the contribution of generative adversarial networks (GANs) \cite{goodfellow2014generative}.
GANs are a framework for learning a generative model.
In a GAN, two neural networks, one for generating synthetic data and the other for discriminating the synthetic data from actual data, are simultaneously trained while competing with each other.
A GAN-based method allows for the generation of data similar to given observations without the domain-dependent knowledge of a target.
A large number of studies have been conducted using GANs for various purposes.
In particular, numerous studies on GANs have been reported in the image domain for tasks such as image super resolution \cite{Ledig2017}, training stabilization \cite{miyato2018spectral}, and domain transformation \cite{Jun2017}.
However, these studies mainly focus on the generation of images, and only a few studies have reported the generation of time-series data \cite{mogren2016crnngan, Yu2017SeqGAN, Dong2018, Esteban2017}.

Recently, we reported that biosignals can be generated using a GAN framework and that the generated signals are effective for data augmentation for biosignal classification \cite{harada2018}.
In \cite{harada2018}, the internal structure of each neural network in a GAN was developed based on a recurrent neural network (RNN) using long short-term memory (LSTM) \cite{hochreiter1997long} for its hidden layers, thereby allowing for the adaptation of the GAN framework to time-series data generation.
Several generative models of biosignals require the domain-dependent knowledge of target biosignals.
In contrast, this method does not require domain-dependent knowledge.
The validity of the biosignal generation method proposed in \cite{harada2018} was qualitatively evaluated using the overall similarity between training and generated data.
The effectiveness for data augmentation was shown via biosignal classification experiments.

However, there were the following limitations in \cite{harada2018}:
\begin{itemize}
\item The GANs should be prepared and independently trained for each class,  resulting in an increase in the number of model parameters in proportion to the number of classes.
\item The generated data have not been evaluated quantitatively.
\item The behavior of the generator is unclear.
\end{itemize}

In this study, we propose a conditional generation method capable of generating multiple classes of time-series data from one model.
The technical highlight of our study is to control the characteristics of the data generated from the proposed method by clarifying the relationship between the input and generated data.
In the proposed method, class labels are simultaneously input to a generator and a discriminator and adapted to conditional generation.
The aim of the proposed method is to reduce training cost and clarify the difference between the classes of training data by training the time-series data of multiple classes with a single model, in contrast to our previous method.
In the experiment, the quality of the generated data is quantitatively evaluated using the similarity between the data generated by the proposed method and the training data.
It is difficult to control the characteristics of the generated data because the input--output relationship in ordinary GANs is unclear.
Therefore, we analyze the input--output relationship of GANs and control the generated data by referring to the analysis result.

The primary contributions of this work are as follows:
\begin{itemize}
\item A conditional method for generating multiple classes of biosignals from a single model is developed. 
\item The performance of the proposed method is quantitatively verified.
\item The behavior of the GAN-based generative model is analyzed to control the characteristics of the data generated by the proposed method.
\end{itemize}

\section{Related Work}
\subsection{Biosignal Generation Models}
Various biosignal generation models have been investigated for a long time \cite{Koski1996, Yamanobe1998, Farina2001, Wendling2000, Mcsharry2003}.
The purpose of such studies is two-fold. One is the understanding of the mechanism of biological systems \cite{Yamanobe1998, Rempe2018, Silva2003}. 
For example, Silva \textit{et al.} \cite{Silva2003} presented their view of the basic mechanisms of the routes to epileptic seizures.
The other is the generation of data for evaluating biosignal processing algorithms \cite{Wendling2000, Mcsharry2003, Farina2001}.
McSharry \textit{et al.} proposed an ECG generation model based on three ordinary differential equations.

Biosignal generation models are categorized into two approaches, i.e., the mathematical model-based approach and machine learning-based approach.
For the mathematical model-based approach, McSharry \textit{et al.} proposed an ECG generation model based on differential equations \cite{Mcsharry2003}.
This model consists of three ordinary differential equations and can control various characteristics of generated signals, such as the interval between waves and the value of P-waves and Q-waves.
Wendling \textit{et al.} proposed a multiple coupled populations model, where each single population model consists of ordinary differential equations \cite{Wendling2000}.
As an example of machine learning-based approaches, Koski \textit{et al.} \cite{Koski1996} proposed an ECG generation model based on hidden Markov models (HMM).
In \cite{Koski1996}, artificial ECG signals were generated using an HMM and two-class classification between normal and pathological ECG was performed using an HMM.
Even though both abovementioned approaches have the possibility of generating high-quality data with characteristics similar to original data, 
they have their advantages and disadvantages. 
On one hand, the mathematical model-based approach can change the characteristics of generated data by adjusting parameters; however, domain-dependent knowledge is required.
On the other hand, the machine learning-based approach does not require domain-dependent knowledge and can therefore be applied to general applications, whereas the model-based approach performs well only if the assumed model structure sufficiently approximates the true data distribution.

The proposed method is a machine learning-based approach.
Our method does not require distribution assumptions because it is based on a GAN, which is a neural-network-based generative model; however, this method generates models with low interpretability.
Therefore, in this study, an analysis was performed from various viewpoints to clarify the behavior of our method.

\subsection{GANs}
\begin{figure}[tb]
  \centering
  \subfloat[GANs for image generation]{
      \includegraphics[width=.5\columnwidth]{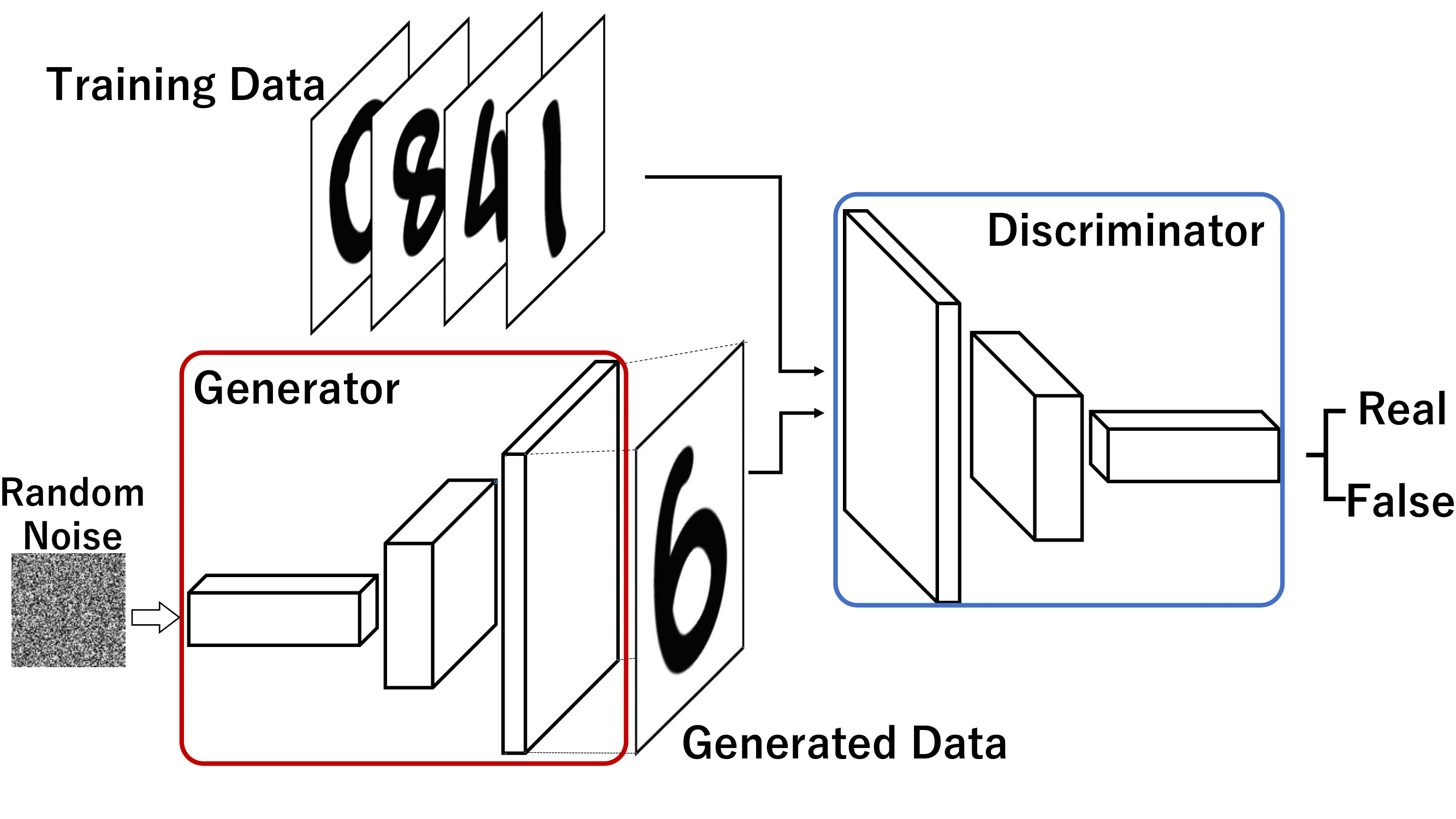}}
  \subfloat[GANs for time-series generation]{
      \includegraphics[width=.5\columnwidth]{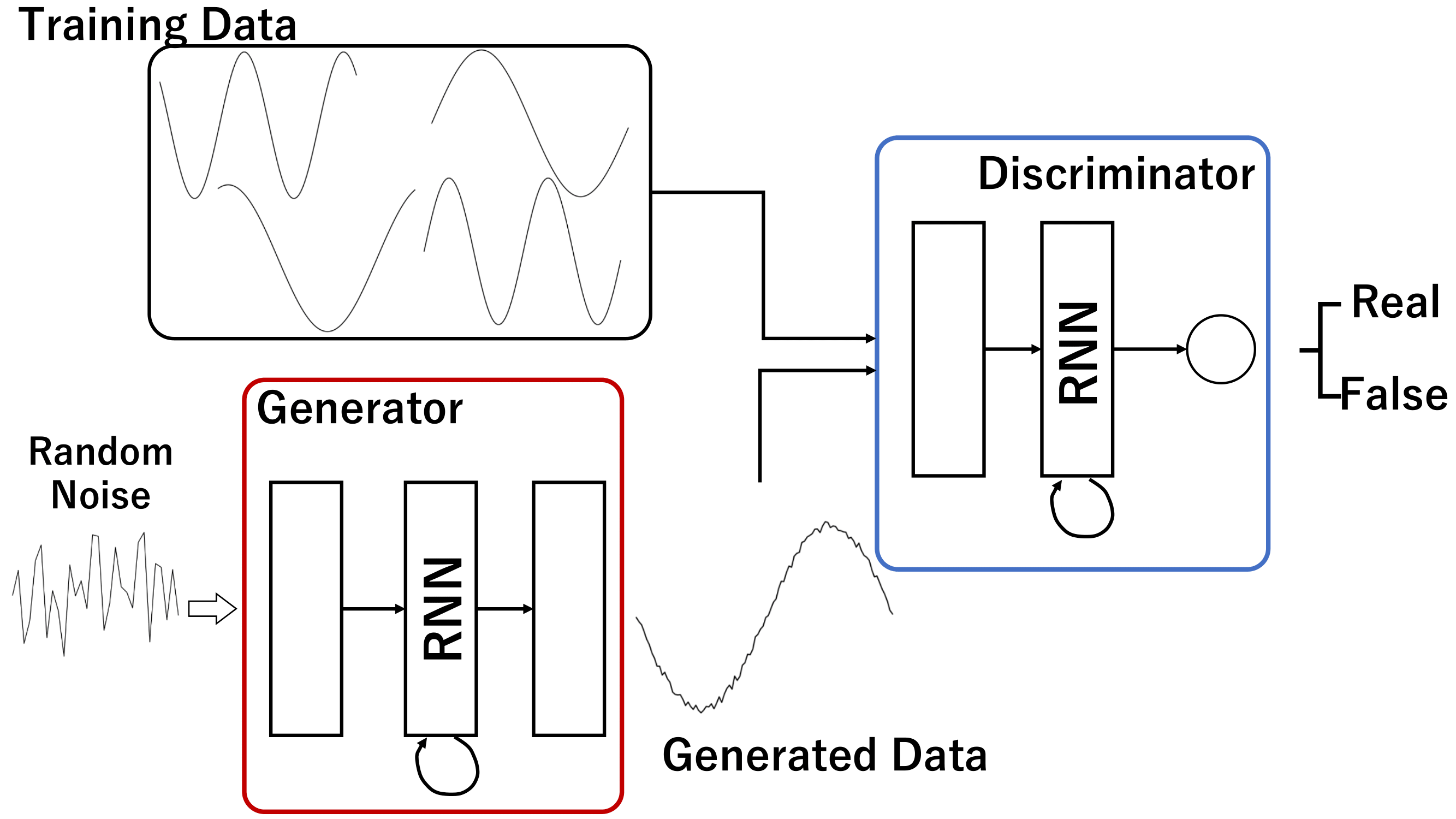}}
    \caption{Overview of the GAN framework.}
  \label{gans_overview}
\end{figure}
GANs are a method for estimating generative models proposed by Goodfellow \textit{et al.} in 2014 \cite{goodfellow2014generative}.
Fig. \ref{gans_overview} shows an overview of the GAN framework.
Using this framework, it is possible to generate data similar to given observations without the domain-dependent knowledge of a target.
GANs have received considerable attention in recent years, particularly in the computer vision community, and various derivatives have been proposed by changing their learning methods and structures.

A GAN consist of two different networks.
One is a generative model referred to as a generator. 
A vector of random numbers is fed into the generator. The generator produces data with the same dimensions as those of training data.
The other network is a discriminative model referred to as a discriminator.
Training data and the data generated by the generator are input to the discriminator.
Then, the network discriminates whether the input came from the training data or generated data.

The generator and discriminator are repeatedly trained in the GAN framework.
Their relationship is frequently compared to that of banknote counterfeiters and police.
The generator learns to generate data that the discriminator classifies as training data.
In contrast, the discriminator learns to discriminate training data and generated data correctly.
As a result, the generator gradually gains the ability to generate data that are similar to but not completely the same as training data.
In other words, the generator learns the mapping from a distribution of random numbers onto the distribution of training data.

In recent years, numerous studies have been performed using GANs.
Researchers have proposed new GANs for various purposes such as the improvement of learning stability, the generation of high-resolution images, and translation to different class images.
For example, using Wasserstein GANs, Arjovsky \textit{et al.} improved learning stability, prevented mode collapse, and provided meaningful learning curves useful for hyperparameter searches.
Moreover, methods of conditional generation using GANs have been reported \cite{Esteban2017, Odena2017, Chen2016}.
These methods achieve conditional generation by considering auxiliary information.

These studies mainly focus on the generation of images, and only a few studies have reported the generation of time-series data \cite{mogren2016crnngan, Yu2017SeqGAN, Dong2018, Esteban2017}.
For example, Yu \textit{et al.} proposed SeqGAN for natural language generation \cite{Yu2017SeqGAN}.
In SeqGAN, the generator and discriminator are constructed based on LSTMs and reinforcement learning is applied to the training of the generator.
Dong \textit{et al.} proposed a music generation method based on convolutional GANs \cite{Dong2018}.
This method consists of multiple generators and discriminators that generate and identify the sound of each track.

Our previous study \cite{harada2018} has demonstrated the effectiveness of data enhancement using biosignals generated from GANs.
In \cite{harada2018}, biosignal generation and biosignal data augmentation were performed by a time-series data generation method based on GANs constructed with LSTM.
However, certain limitations exist in our previous study.
First, the proposed GAN-based method must be trained independently for each class.
Second, the quality of generated data was not quantitatively evaluated.
Finally, the latent variable space was not analyzed.

This study proposes a generation method based on GANs that can select the class of generated data based on the class label. In this method, it is not required to learn models independently in each class.
Furthermore, the behavior of the proposed GAN is grasped through input--output analysis, and the generated data characteristics not considered during learning are controlled using the analysis results.
Control by a class label is the same as existing conditionally generated GANs.
The contribution of this work is to control the characteristics of the generated data not considered during training by analyzing the trained model.

\begin{figure*}[tb]
\hfill
  \begin{minipage}[b]{.9\columnwidth}
    \centering
    \subfloat[Structure of the proposed method.]{\includegraphics[width=1\columnwidth]{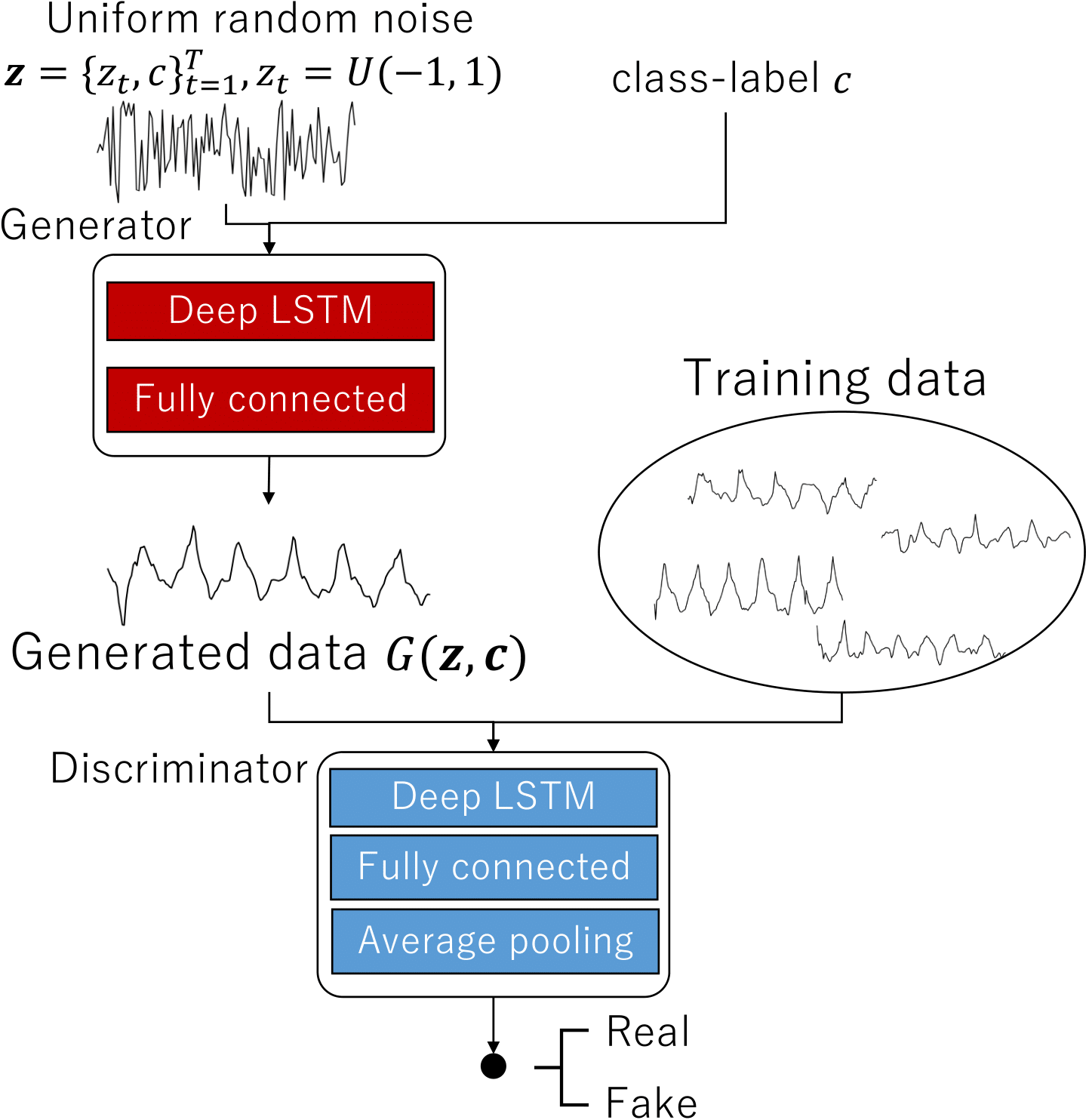}
    \label{fig:gans}}
  \end{minipage}
\hfill
  \begin{minipage}[b]{.9\columnwidth}
    \centering
    \subfloat[Internal structure of the generator.]{\includegraphics[width=0.8\columnwidth]{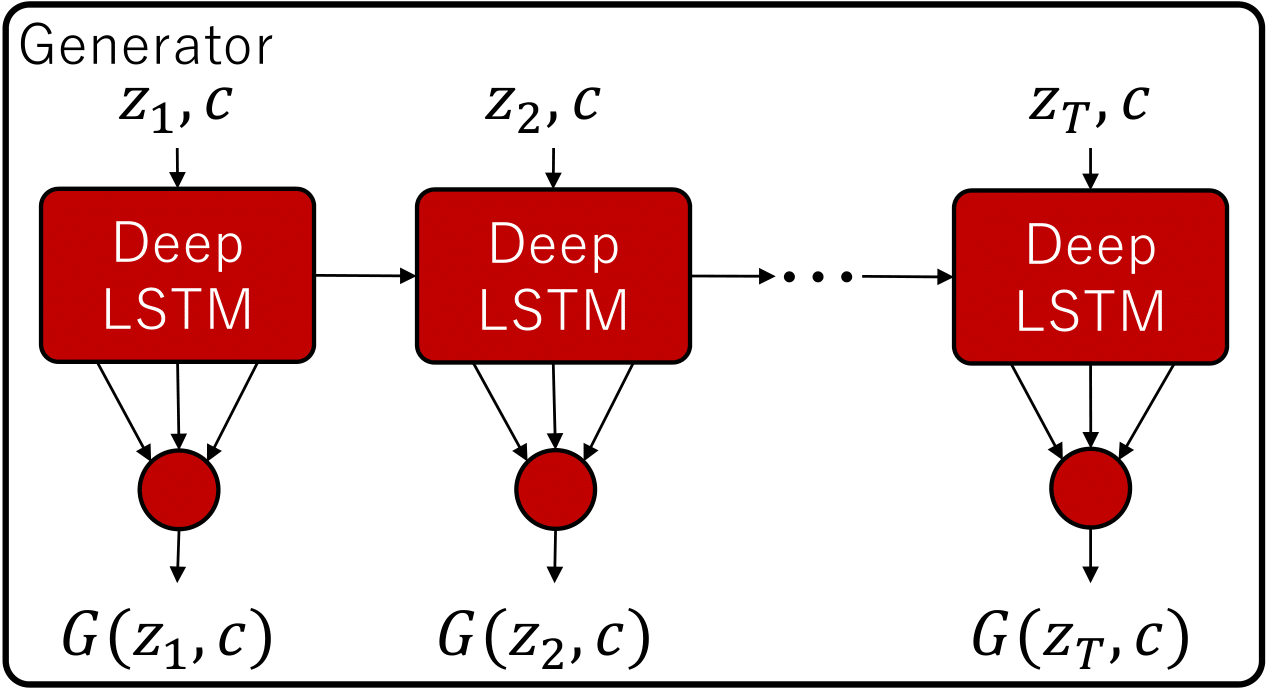}
    \label{fig:gene}}\\
    \subfloat[Internal structure of the discriminator. The triangular node indicates multiplication.]{\includegraphics[width=0.8\columnwidth]{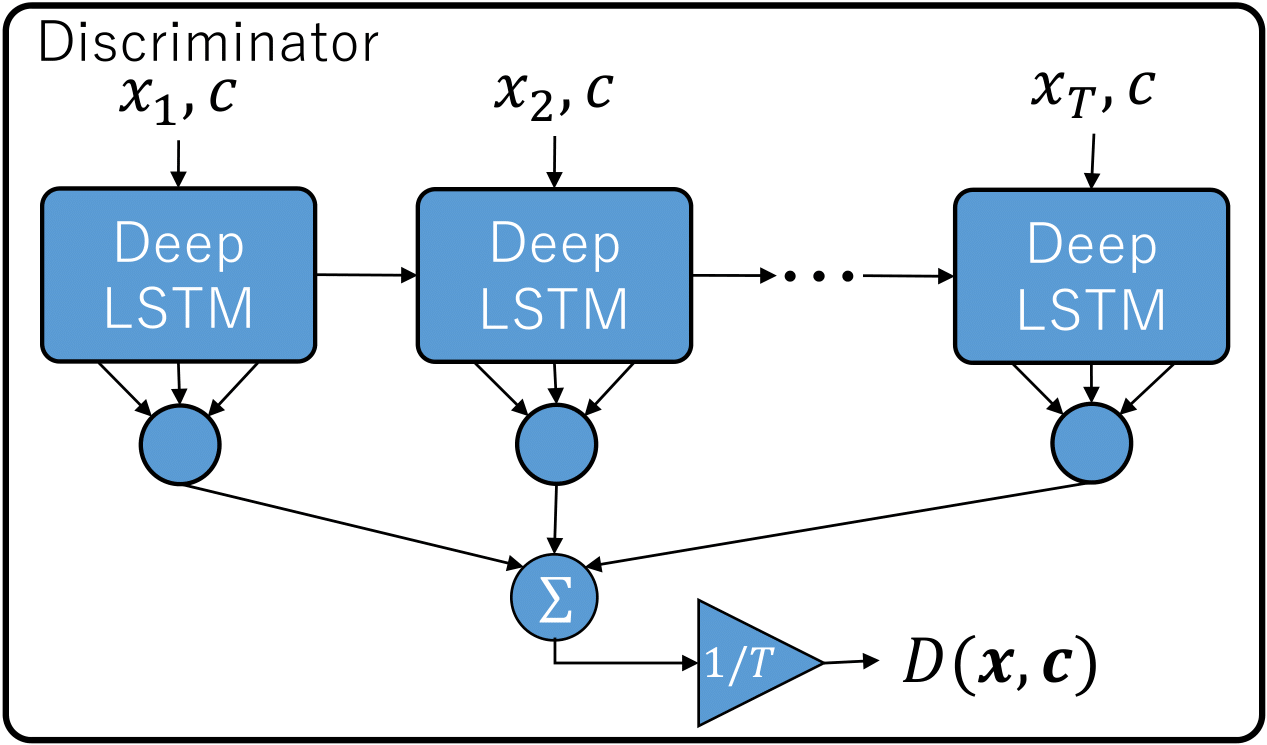}
    \label{fig:disc}}
  \end{minipage}
    \caption{Overview of the proposed method. On one hand, the generator learns to generate data similar to the original biosignal. On the other hand, the discriminator learns to discriminate the data generated by the generator and the original biosignal. By learning these neural networks alternately, the generator can generate data close to the original biosignal.}
    \label{model}
\end{figure*}

\begin{algorithm}[h]                  
    \caption{Training procedure of the proposed method}         
    \label{algo}
    \begin{algorithmic}
    \Require
        \Statex $\bm{X} = \{\bm{x}^{(1)}, \bm{x}^{(2)}, \cdots , \bm{x}^{(M)} \}$: Training dataset
        \Statex $p_z = U(-1, 1)$: Uniform distribution
        \Statex $p_c(k) = \frac{1}{C}, \ k = \{1, \cdots C\}$: Discrete uniform distribution
        \Statex $C$: Number of classes
    \Ensure
      \Statex $\bm{\theta_{d}}$: Weights of the discriminator $D$
        \Statex $\bm{\theta_{g}}$: Weights of the generator $G$
    
    \For{number of training iterations}
    \For{number of unrolling $+ 1$}
    \begin{itemize}
    \setlength{\leftskip}{0.5cm}
    \item Sample minibatch of $m$ noise sequence samples $\{\bm{z}^{(1)}, \bm{z}^{(2)}, \cdots , \bm{z}^{(m)}\}$ from noise prior $p_z$
    \item Randomly generate $m$ class-label sequences $\{ \bm{c}_z^{(1)}, \bm{c}_z^{(2)}, \cdots , \bm{c}_z^{(m)} \}$ from $p_c$
    \item Sample minibatch of $m$ examples $\{ \bm{x}^{(1)}, \bm{x}^{(2)}, \cdots , \bm{x}^{(m)} \}$ from dataset
    \item Sample $m$ class-label sequence $\{ \bm{c}_x^{(1)}, \bm{c}_x^{(2)}, \cdots , \bm{c}_x^{(m)} \}$ corresponding to $m$ sampled $\bm{x}$,
    \item Update the discriminator by ascending its stochastic gradient:
    \begin{align*}
    \hspace{.5cm}
    \scalebox{0.8}{$\displaystyle
    \nabla_{\bm{\theta_d}} \frac{1}{m}\sum_{i=1}^{m} \left[\log D \left(\bm{x}^{(i)}, \bm{c}_x^{(i)} \right) + \log \left(1 -
    D \left(G \left(\bm{z}^{(i)}, \bm{c}_z^{(i)} \right) \right) \right) \right].
    $} \notag
    \end{align*}
    \end{itemize}
    \hspace{.5cm}
    \If{first update at this iteration}
    \begin{itemize}
    \setlength{\leftskip}{1.0cm}
    \item Save weights of the discriminator $\bm{\theta_{d}}^{(1)}$
    \end{itemize}
    \EndIf
    
    \EndFor
    \begin{itemize}
    \item Sample minibatch of $m$ random sequence samples $\{\bm{z}^{(1)}, \bm{z}^{(2)}, \cdots , \bm{z}^{(m)}\}$ from noise prior $p_z$
    \item Randomly generate $m$ class-label sequences $\{ \bm{c}_z^{(1)}, \bm{c}_z^{(2)}, \cdots , \bm{c}_z^{(m)} \}$ from $p_c$
    \item Update the generator by descending its stochastic gradient:
    \begin{align*}
    \scalebox{0.8}{$\displaystyle
    \nabla_{\bm{\theta_g}} \frac{1}{m}\sum_{i=1}^{m}\log \left(1 - D \left(G \left(\bm{z}^{(i)}, \bm{c}_z^{(i)} \right) \right) \right).
    $} \notag
    \end{align*}
    \item Load weights of the discriminator $\bm{\theta_{d}}^{(1)}$
    \end{itemize}
    \EndFor
    \end{algorithmic}
    \end{algorithm}

\section{Time-Series Data Generation Method}
Fig. \ref{model} shows the structure of the proposed method.
Based on the GAN framework, the proposed method consists of generator $G$ and discriminator $D$.
The proposed method is composed of an RNN based on LSTMs to adapt to time-series data, whereas most existing GANs are constructed based on convolutional neural networks.

Generator $G$ consists of a deep LSTM layer and a fully connected layer.
The deep LSTM layer has $L$ hidden layers and $U$ LSTM units in each hidden layer.
The fully connected layer has a sigmoid function as the activation function.
The generator receives a latent variable sequence, $\bm{z} = \{\it{z}_t\}_{t=1}^{T}$, and an additional information sequence, $\bm{c} = \{\it{c}_t\}_{t=1}^{T}$ ($T$ is the length of the training data), as input data.
More specifically, generator $G$ receives a sequence with $\it{z}_t$ and $\it{c}_t$ simultaneously at each time point $t$, where $\it{z}_t$ and $\it{c}_t$ are combined and input as a vector.
At each time point $t$, $\it{z}_t$ is independently sampled from a uniform random distribution, $U(-1, 1)$.
In addition, the class-label sequence, $\bm{c}_z = \{ c_t|c_{t+1}=c_t\}$, is a constant sequence where the initial value is sampled from a discrete uniform distribution, $p_c(c_1) = \frac{1}{C}$, for each latent variable $\bm{z}$, where $C$ is the number of classes.
The sequence of the output from the fully connected layer is treated as the conditionally generated time-series data, $G(\bm{z}, \bm{c}) \in \mathbb{R}^T$.

Discriminator $D$ consists of a deep LSTM layer, a fully connected layer, and an average pooling layer.
The deep LSTM layer in the discriminator has the same number of hidden layers and LSTM units as the generator in each hidden layer, and the fully connected layer has a sigmoid function as the activation function.
The average pooling layer outputs a scalar by averaging the input over its dimensions.
As with generator $G$, discriminator $D$ also receives a sequence with $\it{x}_t$ and $\it{c}_t$ simultaneously at each time point $t$.
The input data, $\bm{x}$, are sampled from training data or the uniform random distribution, $U(-1, 1)$.
If $\bm{x}$ is sampled from training data, the $\bm{c}$ corresponding to that $\bm{x}$ is a class label of that $\bm{x}$.
If $\bm{x}$ is $G(\bm{z})$ generated from generator $G$, the $\bm{c}$ corresponding to that $\bm{x}$ is sampled from the discrete uniform distribution, $p_c(k)$, where $z$ is sampled from $U(-1, 1)$.
Given an input sequence, $\bm{x}$, and $\bm{c}$, the output of discriminator $D(\bm{x}, \bm{c})$ is a scalar value representing the probability that $\bm{x}$ came from the training data.

In the training, $D$ and $G$ play the minimax game with the evaluation function defined as
\begin{align}
\begin{aligned}
  \min_G \max_D V(G, D) &= \mathbb{E}_{\bm{x} \sim p_{\rm{data}}(\bm{x})} \left[ \log D \left( \bm{x}, \bm{c} \right) \right] \\
&+ \mathbb{E}_{\bm{z} \sim p_{z}(\bm{z})} \left[ 1 - \log D \left( G \left( \bm{z}, \bm{c} \right) \right) \right],
\end{aligned}
\end{align}
where $p_{\rm{data}}(\bm{x})$ and $p_{z}(\bm{z})$ are distributions of the training data and $\bm{z}$, respectively.
The training procedure is shown in Algorithm \ref{algo}.
The gradients with respect to the weights of the networks are calculated using the backpropagation-through-time method.
Each minibatch for the training of $D$ contains the same number of $\bm{x}$ and $\bm{z}$, and there is no specification of the class label ratio of the minibatch used to train $D$ and $G$.

The weights are updated based on an unrolled GAN, which is an updating rule proposed by Metz \textit{et al.} \cite{UNROLL}.
Using this updating rule, we can avoid mode collapse and prevent the generator from generating biased data.

\section{Biosignal Generation Experiment}
Biosignal generation experiments were conducted using three real-world biosignal datasets to quantitatively evaluate the biosignal generation of multiple classes using the proposed method.
In this experiment, the data generated by the proposed method were evaluated qualitatively and quantitatively.
First, the generated biosignals were qualitatively evaluated by comparing them with the actual biosignals.
Then, the similarity between the generated data and training data was computed to evaluate the quality of the data.

\subsection{Datasets}
Three real-world biosignal datasets from The UEA \& UCR Time Series Classification Repository \cite{bagnall2016} were used in this study. 
All datasets were normalized in a range 0 to 1.
The details of these datasets are as follows:
The first dataset is an ECG dataset referred to as ``ECG 200;'' this dataset was created by Olszewski \cite{ECG200}.
It consists of $200$ samples of an ECG series.
Each series traces the electric activity during one heartbeat, and the length of each series is $96$.
Out of $200$ samples, $133$ are labeled as normal and the remaining $67$ are myocardial infarctions (or abnormal).
In this paper, the normal and abnormal classes of the ECG200 dataset are referred to as class 1 and class 2, respectively.
We randomly extracted $160$ samples for training data from the ECG200 dataset.
The second dataset is an ECG dataset referred to as ``TwoLeadECG,'' which was collected and added to the repository by Keogh \cite{UCRArchive}.
This dataset consists of $1{,}162$ samples of an ECG series, and the length of each series is $82$ (each time series reflects one heartbeat).
In the TwoLeadECG dataset, two different leads of the ECG are considered, and each signal originates from one of these two leads.
Out of $1{,}162$ samples, $581$ are labeled as class $1$ and the remaining $581$ are class $2$.
The aim of the TwoLeadECG dataset is to distinguish between the signals originating from each lead.
In this study, the classes of the TwoLeadECG dataset are referred to as class 1 and class 2.
We randomly extracted $498$ samples of each class as training data.
The third dataset is an EEG dataset referred to as ``Epileptic Seizure Recognition Data Set'' \cite{PhysRevE.64.061907}. 
This dataset contains $9{,}200$ normal series and $2{,}300$ abnormal series recorded during epileptic seizure.
We randomly sampled $9{,}200$ for training data from entire data in the EEG dataset.
The length of each series was $178$.

\subsection{Experimental Setup}
In the experiments of each dataset, the number of LSTM units of the generator and discriminator were $400$, and the number of LSTM layers of the generator and the discriminator were $4$.
The Adam optimizer \cite{adam} with an initial learning rate of $0.0001$ was used for weight updating.
The number of training epochs was set to be $10{,}000$.

\begin{figure*}[tbp]
    \centering
    \subfloat[ECG200 dataset]{
        \includegraphics[width=0.66\columnwidth]{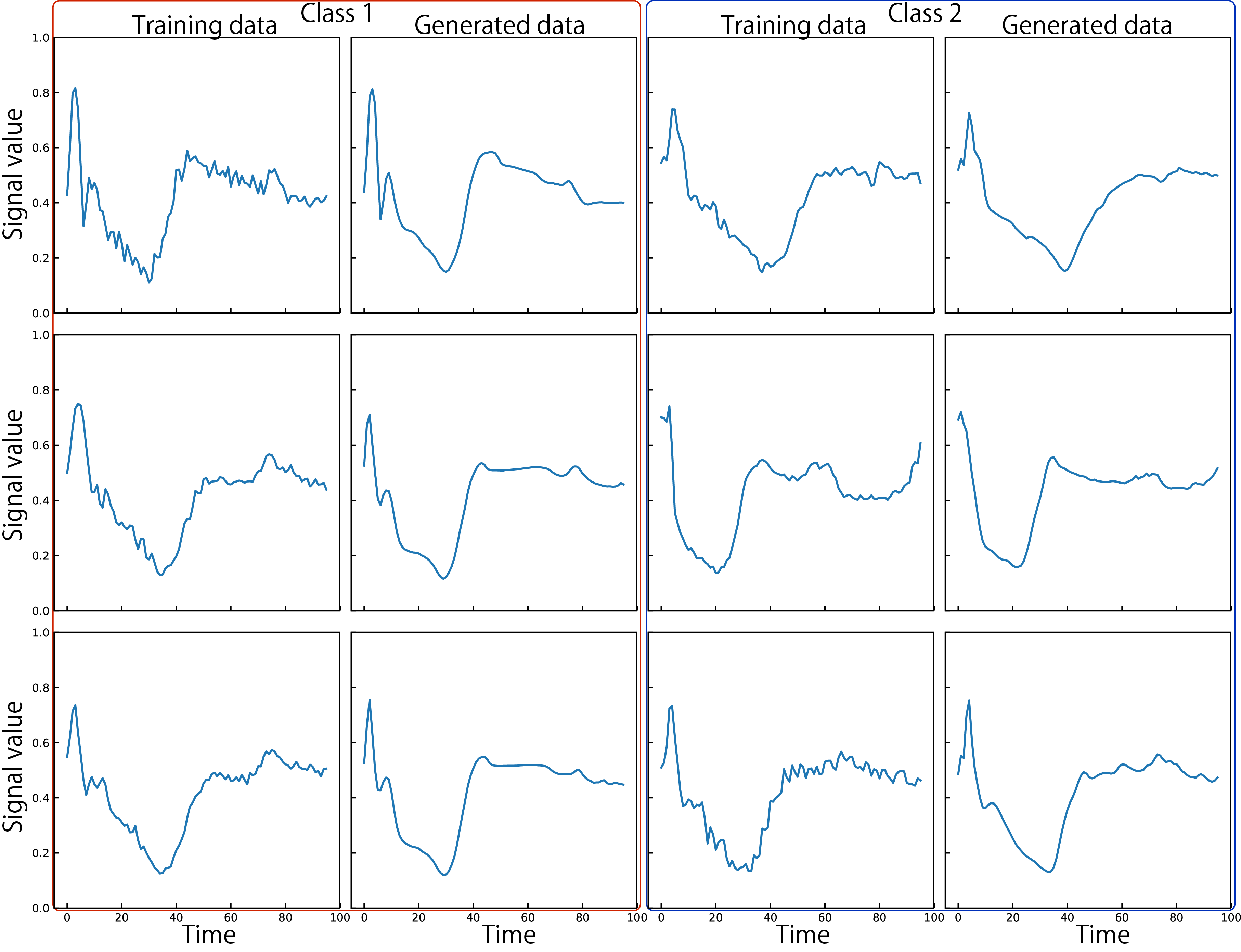}}
    \subfloat[TwoLeadECG dataset]{
        \includegraphics[width=0.66\columnwidth]{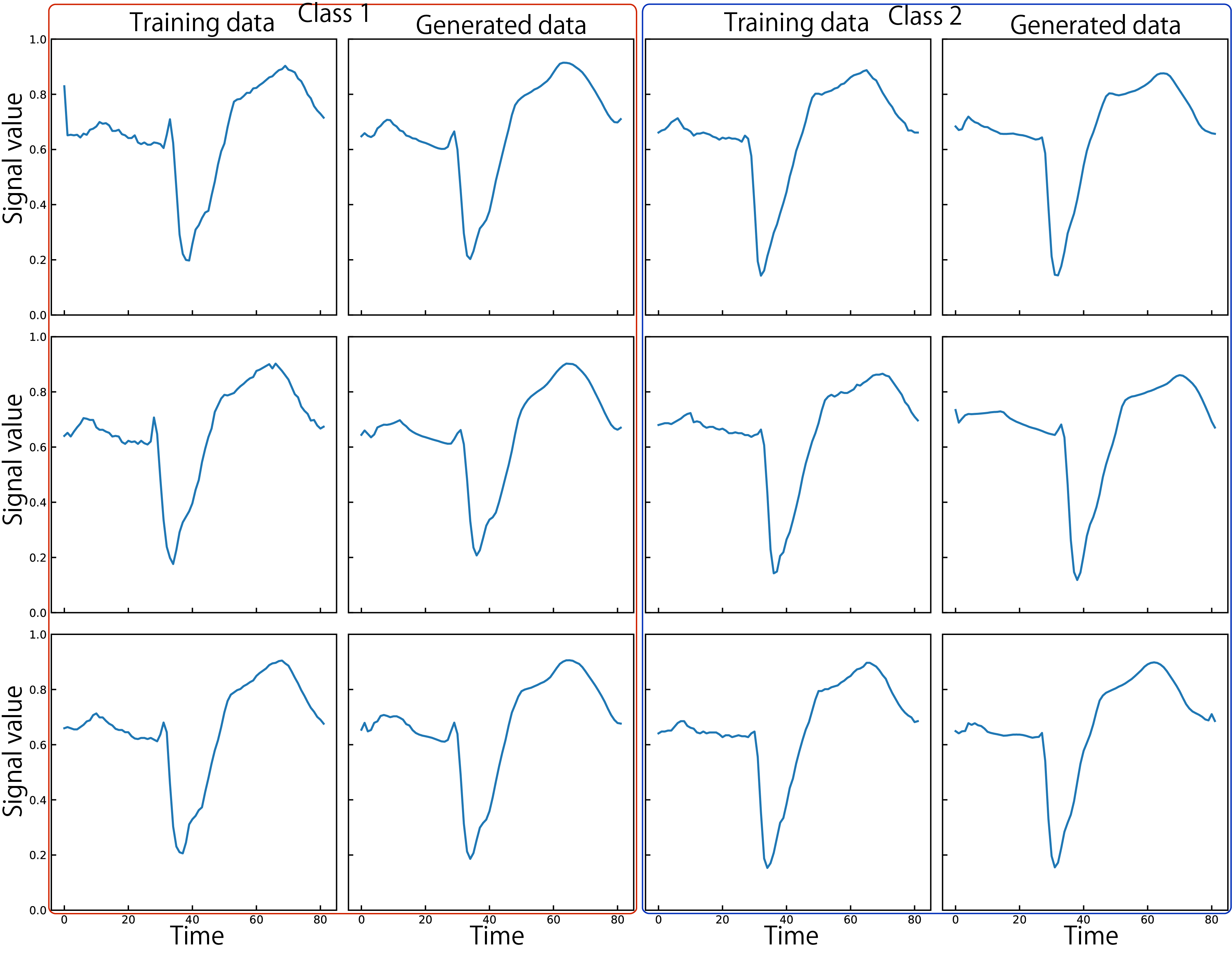}}
    \subfloat[EEG dataset]{
        \includegraphics[width=0.66\columnwidth]{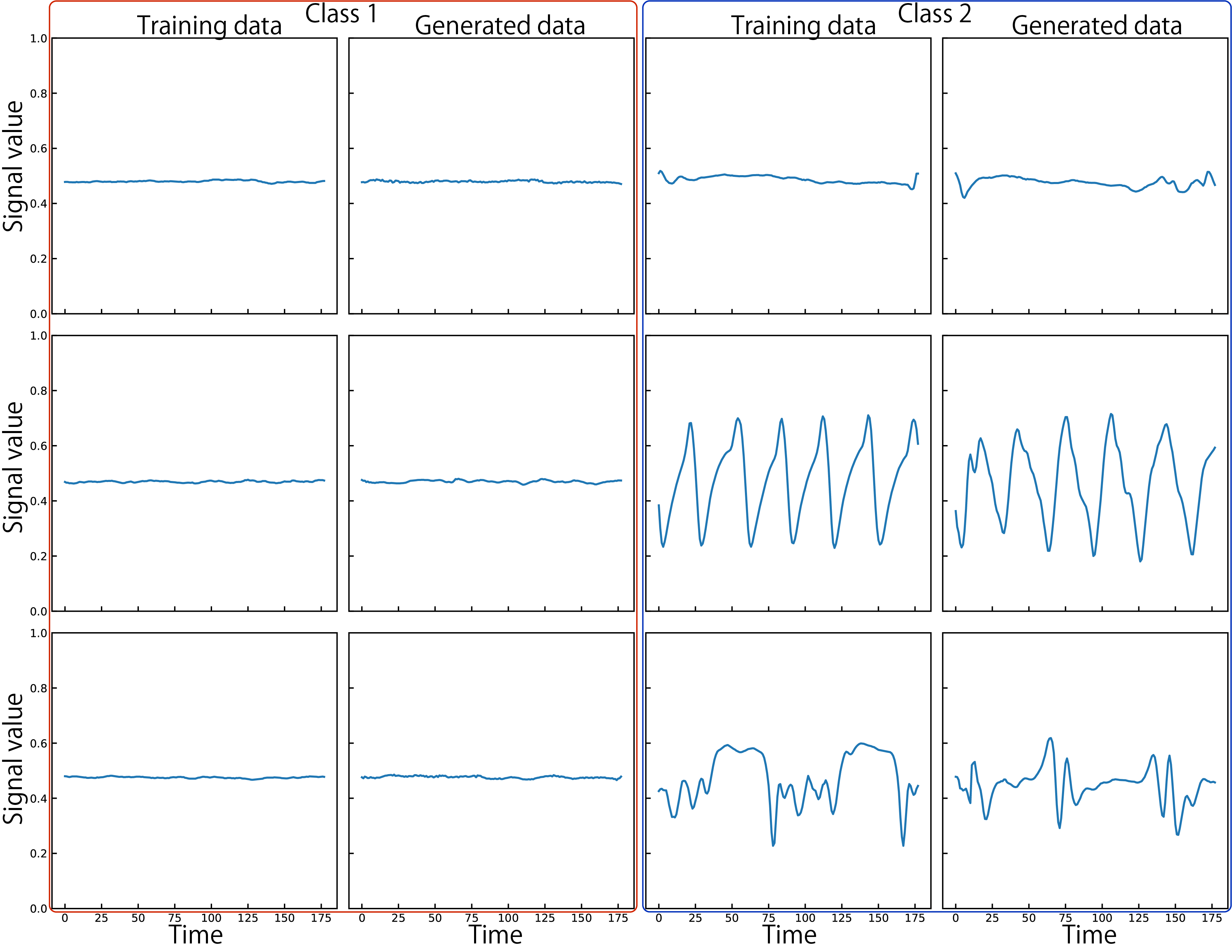}}
    \caption{Example of the original and generated signals. Three medoids obtained by $k$-medoids clustering ($k=3$) from the original dataset are shown as the original signal examples. The signal most similar to each original signal example was selected from the data generated using the proposed method.}
    \label{gene_data}
\end{figure*}

The similarity of the quantitative evaluation was computed by utilizing dynamic time warping (DTW) \cite{Bellman1959, Sakoe1978}. 
The similarity obtained using DTW can be calculated as follows:
Given two time-series data $Q$ and $C$, the DTW distance is computed by finding the best alignment between them.
First, to align the two time-series data, an $n \times m$ matrix is constructed, whose $(i, j)$ element is equal to $(q_i -c_j)$, where $q_i$ and $c_j$ are the points of time-series data $Q$ and $C$, respectively.
An alignment between the two time-series data is represented by a warping path, $W = w_1, w_2, \cdots, w_k, \cdots, w_K$, where $w_k$ is the index of the matrix.
Warping path $W$ starts at the bottom-left corner and ends at the top-right corner of the matrix.
The best alignment is given by a warping path through the matrix that minimizes the total cost of aligning its points, and the corresponding minimum total cost is termed as the DTW distance.
The DTW distance is calculated as
\begin{align}
\scalebox{0.9}{$
\displaystyle
    DTW(Q, C) = \argmin_{W = w_1, \cdots , w_k, \cdots , w_K} \sqrt{\sum_{k=1, w_k=(i, j)}^{K}(q_i - c_j)^2}.
$}
\end{align}

In addition, the quantitative evaluation compares the results obtained using the proposed method and the four existing data generation methods.
The first method adds a noise sequence to the training data.
We generated the noise sequence at each time-point $t$ from a Gaussian distribution with zero mean and standard deviation $\sigma$ calculated across all training data.
New samples can be generated as
\begin{align}
\displaystyle
{\bm{x}^{(i)}}'= \bm{x}^{(i)} + \gamma \bm{X},\ \bm{X} \sim N(0, \sigma^{2}),
\end{align}
where $\bm{x}^{(i)}$ is the $i$-th sample of the training data, and $\gamma$ is a constant value.
In our experiments, $\gamma$ was set to be $0.5$.
The second method generates new data by interpolating between the training data of the same class label.
New samples can be synthesized as follows:
\begin{align}
\displaystyle
{\bm{x}^{(i)}}'= (1 - \lambda) \bm{x}^{(i)} + \lambda \bm{x}^{(j)},
\end{align}
where $\bm{x}^{(j)}$ denotes the training data most similar to $\bm{x}^{(i)}$ and $\lambda$ is a coefficient related to interpolation in a range of $[ 0, 1]$.
The similarity between training data is calculated by Euclidean distance.
In our experiments, we used $\lambda = [0.1, 0.9]$.
The third method extrapolates between the training data of the same class label to generate new data.
New samples are synthesized as
\begin{align}
\displaystyle
{\bm{x}^{(i)}}'= (1 + \lambda) \bm{x}^{(i)} - \lambda \bm{x}^{(j)},
\end{align}
where $\bm{x}^{(j)}$ denotes the training data most similar to $\bm{x}^{(i)}$, and $\lambda$ is coefficient related to interpolation in a range of $[ 0, \infty )$.
The similarity between training data is calculated by Euclidean distance.
In our experiments, we used $\lambda = [0.1, 0.9]$.
The final method uses an HMM to generate data. Each state of the HMM was constructed with a Gaussian mixture distribution.
The parameters of the HMM were estimated using the Baum--Welch algorithm, and the number of states of the HMM was determined based on Akaike's information criterion.

\subsection{Generation Results}
Fig. \ref{gene_data} shows an example of the original data and the data generated using the proposed method.
Figs. \ref{gene_data}(a), (b), and (c) are examples for the ECG200, TwoLeadECG, EEG datasets, respectively.
The left side of each figure is an example of class 1, and the other side is an example of class 2.
In each figure, three medoids obtained by $k$-medoids clustering ($k=3$) are shown as the original signal examples.
For the generated signal examples, a sequence most similar to each original signal example is selected based on the DTW distance.

\subsection{Quality Evaluation}
The average similarity between the original and generated data was computed to evaluate the quality of the data generated using the proposed method. Here, the original data were from the dataset used for training the proposed method, and
the average similarity among the original data was used as a baseline for evaluation.
In the evaluation procedure, first, the same amount of data was selected from each data group, and then, the similarities of all combinations of data for evaluation were calculated by brute force. The average DTW distance and standard deviation were used as the evaluation result.
A small average DTW distance is a good result because this value indicates the dissimilarity between the target data and original data.
However, if this value is extremely small, it implies that the target data are the same as the original data, which is a worse result.

\begin{figure*}[tbp]
    \centering
    \subfloat[Class 1 of ECG200 dataset]{
        \includegraphics[width=0.66\columnwidth]{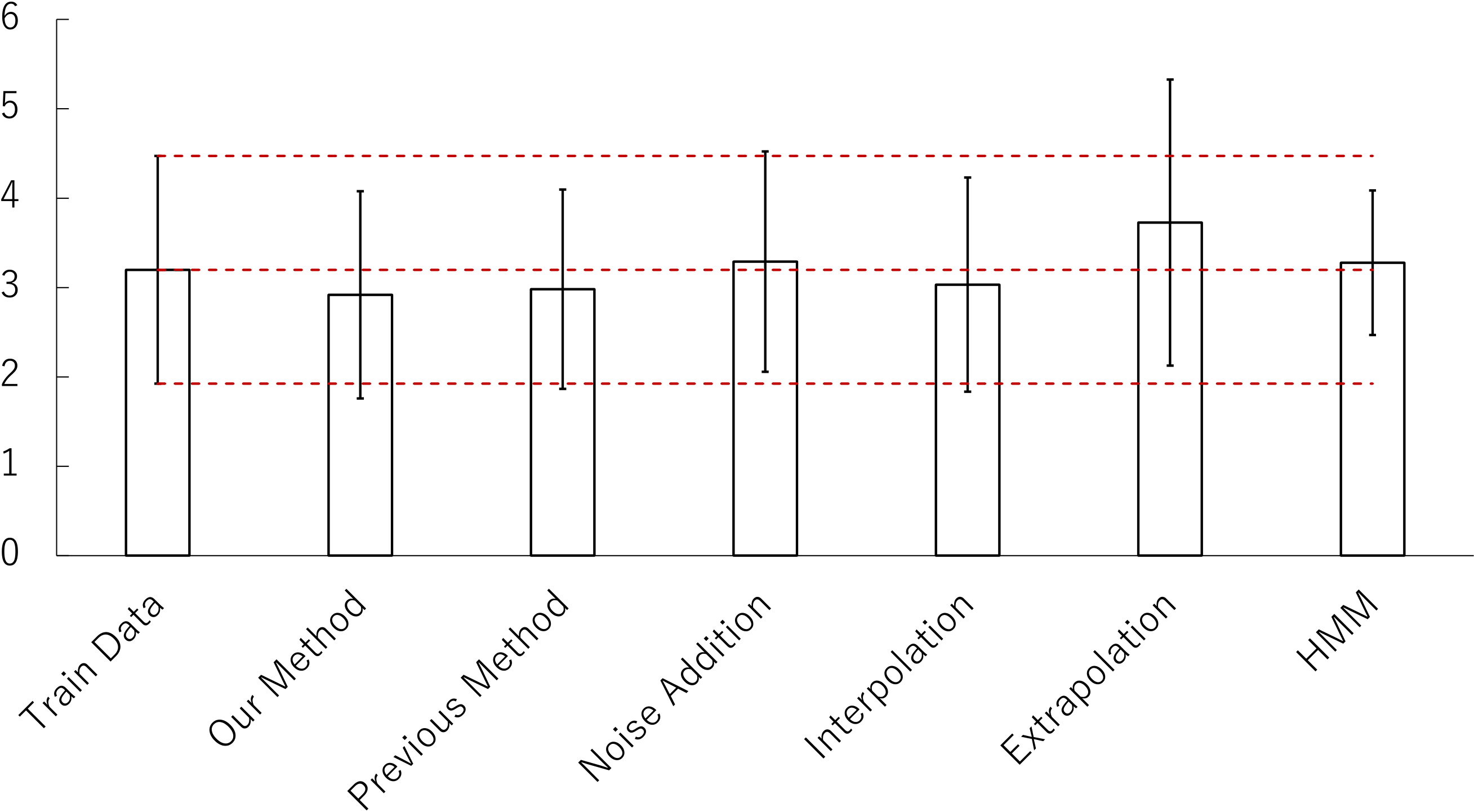}}
    \subfloat[Class 1 of TwoLeadECG dataset]{
        \includegraphics[width=0.66\columnwidth]{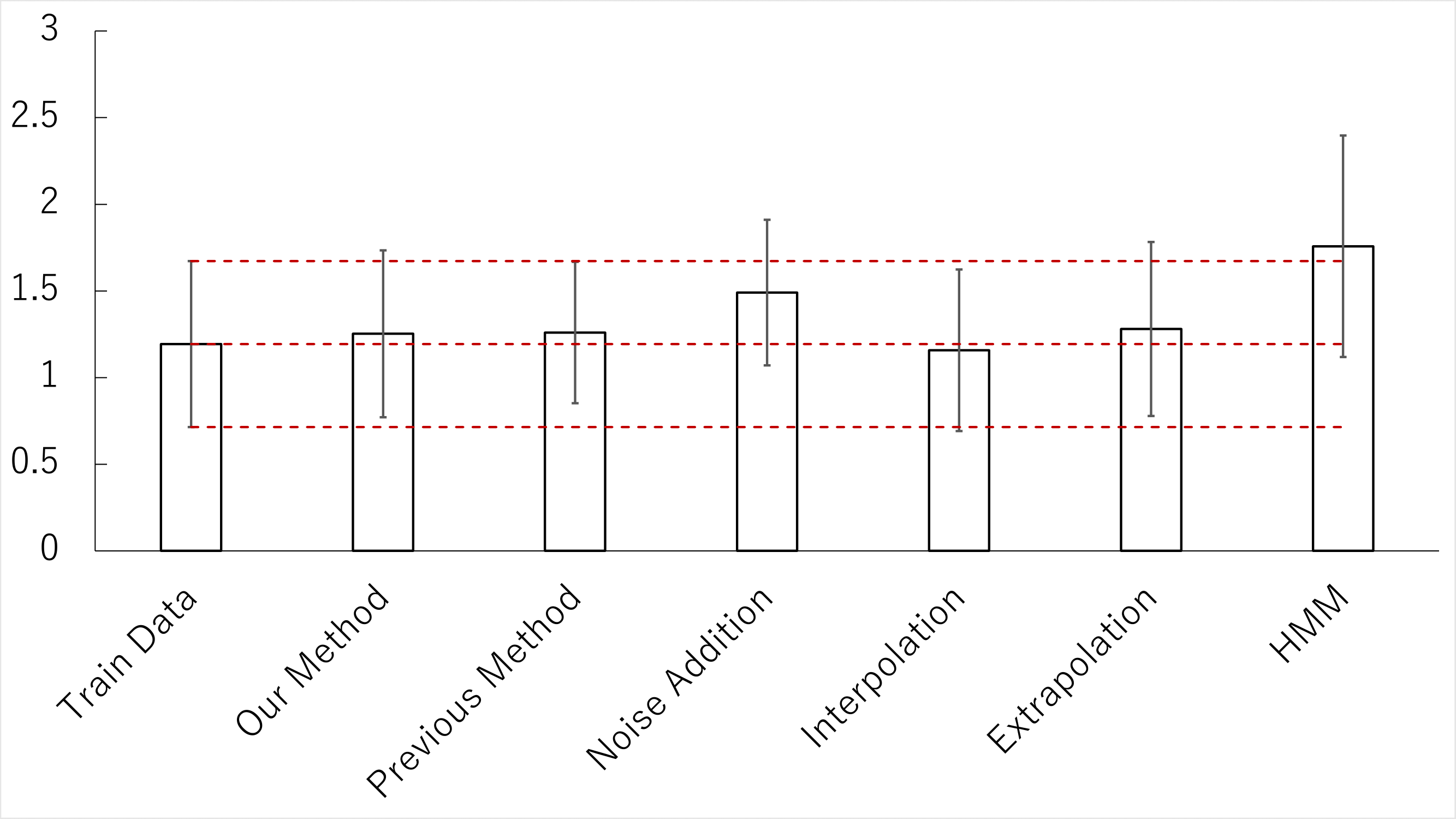}}
    \subfloat[Class 1 of EEG dataset]{
        \includegraphics[width=0.66\columnwidth]{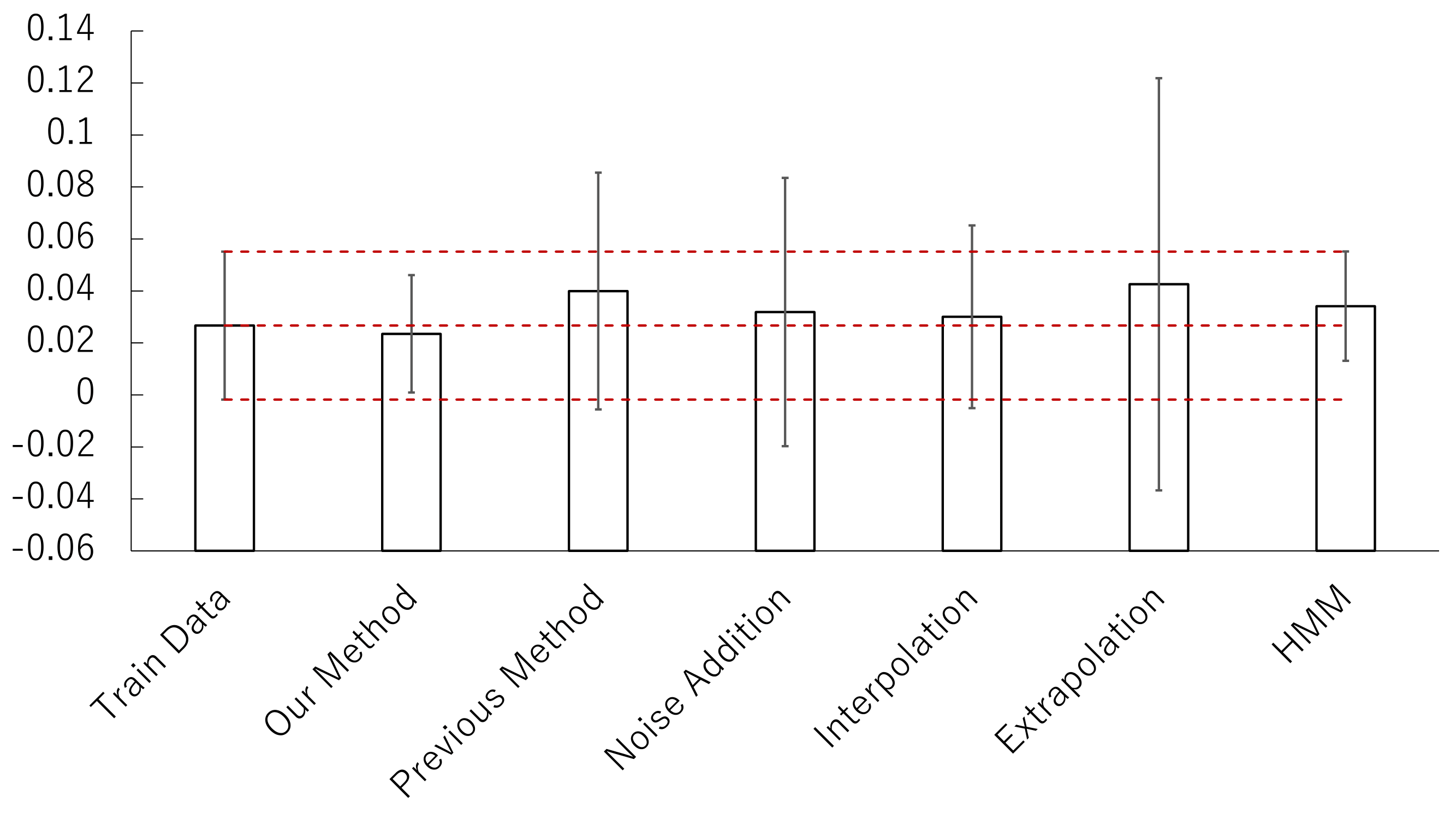}}\\
    
    \subfloat[Class 2 of ECG200 dataset]{
        \includegraphics[width=0.66\columnwidth]{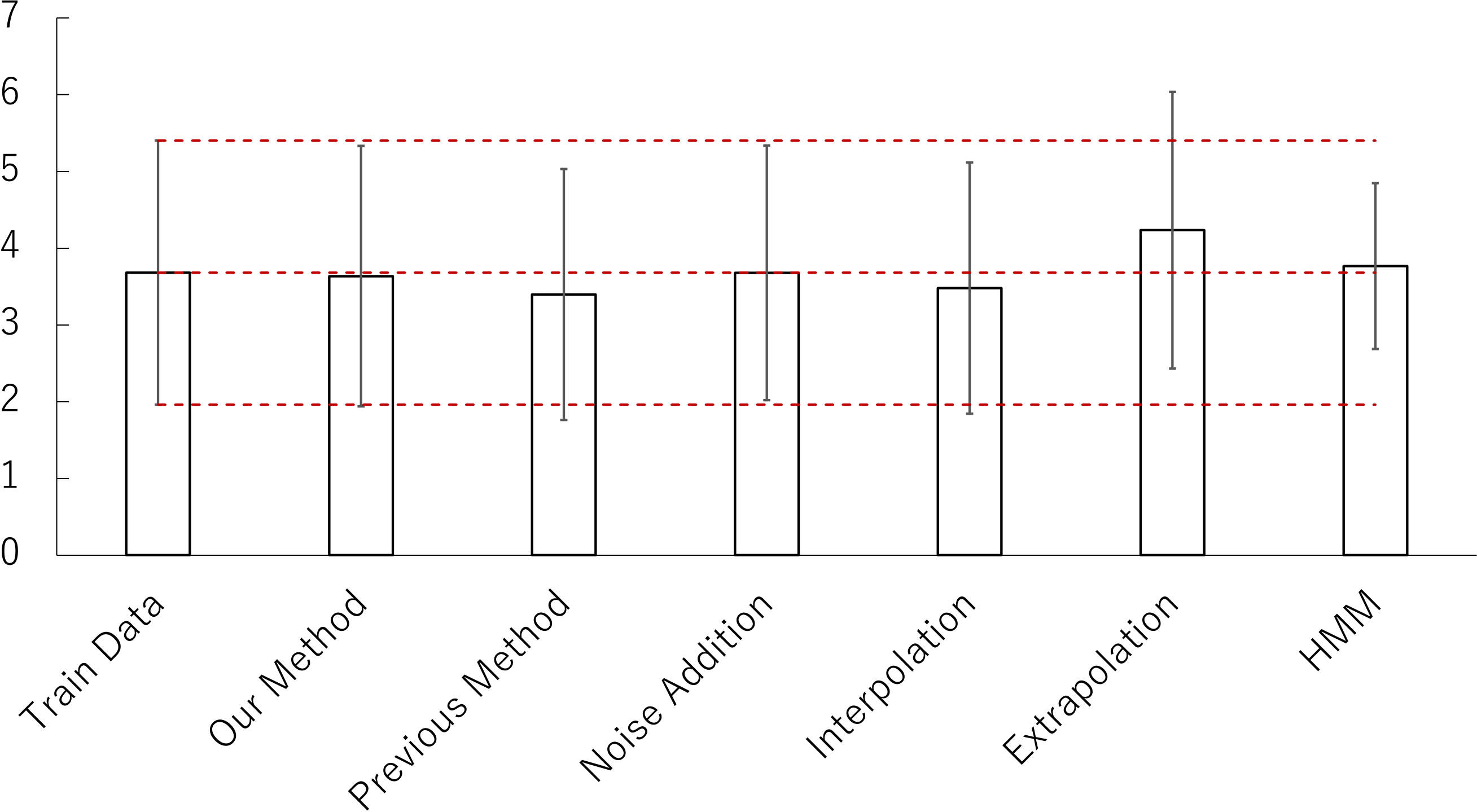}}
    \subfloat[Class 2 of TwoLeadECG dataset]{
        \includegraphics[width=0.66\columnwidth]{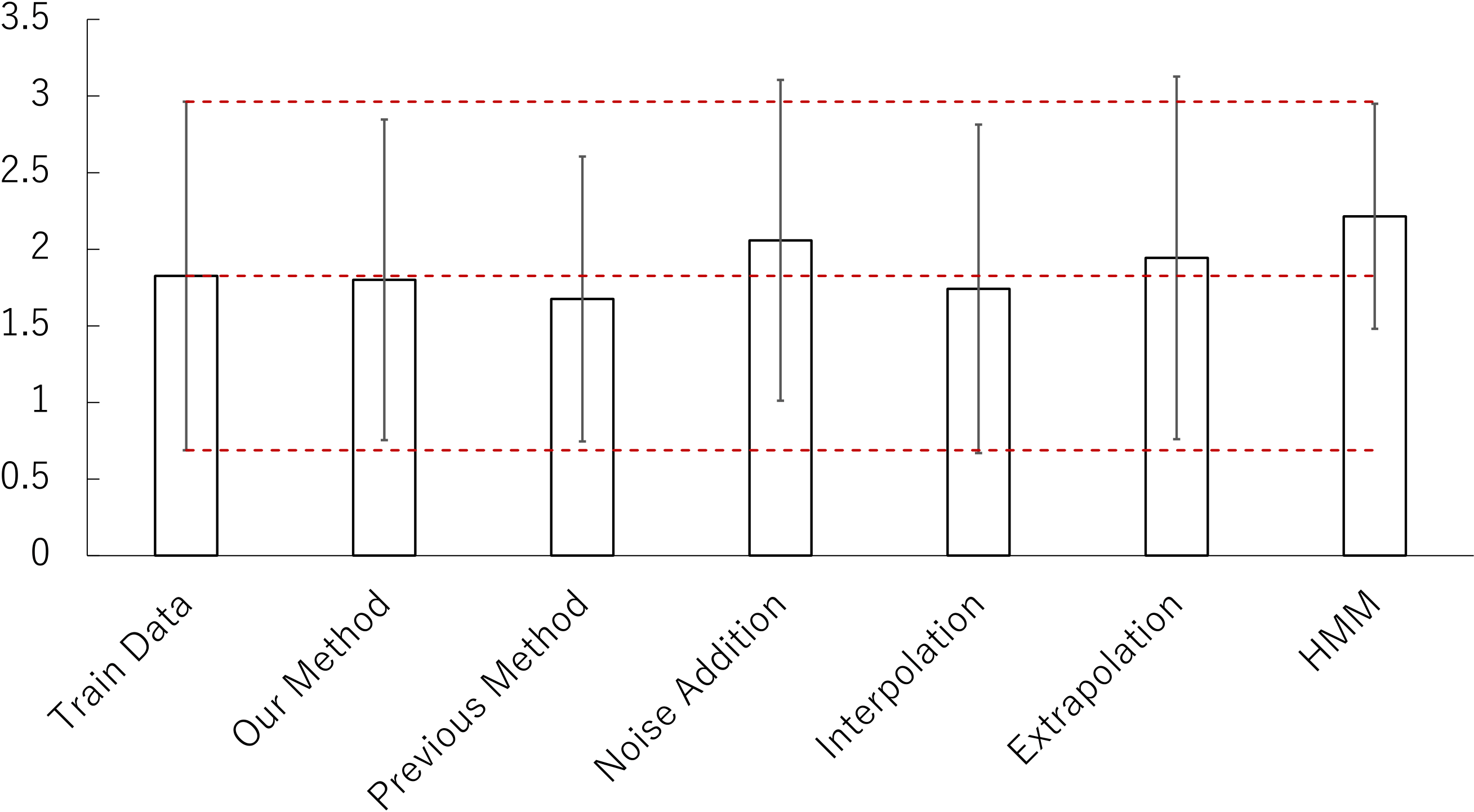}}
    \subfloat[Class 2 of EEG dataset]{
        \includegraphics[width=0.66\columnwidth]{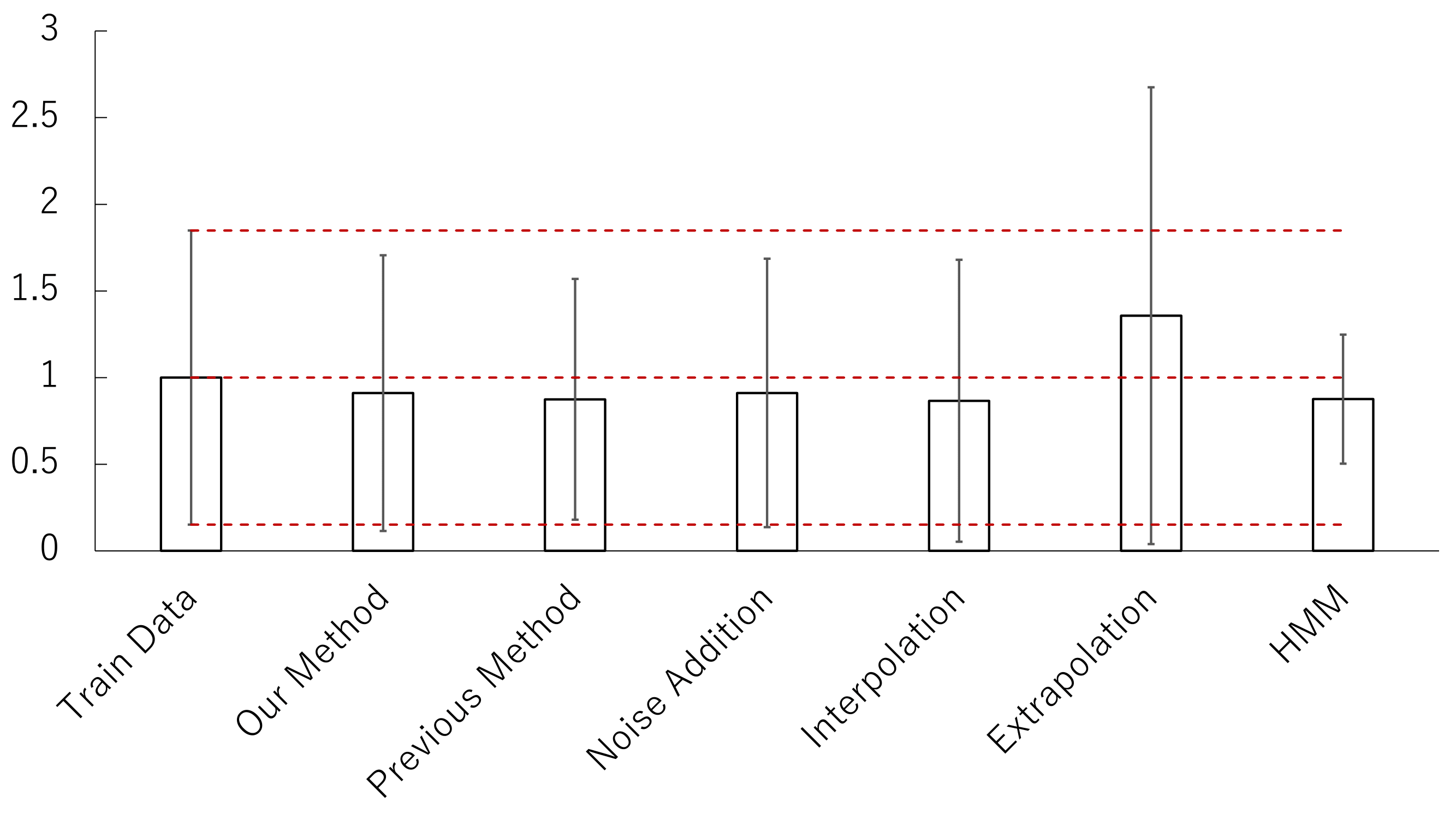}} \\
    \caption{Similarity of the data generated by each data augmentation method. Each bar in the graph shows the average value and standard deviation. The red dashed line indicates the accuracy of the classifier when data augmentation    is not applied.}
    \label{evaluate}
\end{figure*}

Fig. \ref{evaluate} shows the quality of the data generated by each data augmentation method.
Figs. \ref{evaluate}(a) and (d) show the evaluated result of class 1 and class 2 of the ECG200 dataset, respectively.
Figs. \ref{evaluate}(b) and (e) show the evaluation result of class 1 and class 2 of the TwoLeadECG dataset, respectively.
Figs. \ref{evaluate}(c) and (f) show the evaluation result of class 1 and class 2 of the EEG dataset, respectively.
Each bar indicates the average DTW distance among the original data and between the original data and target data generated using each data augmentation method.
The horizontal axis labels indicate the evaluated data, the vertical axis represents the average similarity obtained using the DTW distance, and the error bar indicates the standard deviation of these similarities.

\section{Analysis of the Input--Output Relationship}
Two experiments were performed to analyze the input--output relationship of the generator of the proposed method and confirm controllability. One was to analyze class labels and the other to analyze the latent variable space.
Class-label analysis was performed to evaluate the discrimination between the classes of the generated data.
Latent variable space analysis was performed to clarify the relationship between the input data as a latent variable and the characteristics of the generated data.
Furthermore, the characteristics of the generated data were controlled using the results of this analysis.

\subsection{Analysis of Effect of Class Labels }
Class labels were interpolated to verify whether it is possible to generate data that reflect the features of each class according to a given class label.
An input class label, $\bm{c}$, was obtained by linear interpolation between the original class labels.
If $\bm{c}$ is close to a certain class label, the generated data strongly reflect the characteristics of the training data of the class.
In addition, the difference between the data of each class was confirmed using the transition of the data generated by the proposed method.

The generator of the proposed method was given a fixed random sequence, $\bm{z}$, and class label $\bm{c} = \{ \bm{c}^{(1)}, \cdots , \bm{c}^{(100)} \}$. The data were divided into 100 class 1 and class 2 samples.
By comparing the data generated with class label $\bm{c}^{(i)}$ and the average value at each time point of the training data of each class, it is demonstrated that the generated data reflect the features of each class of training data.

\begin{figure*}[tbp]
    \centering
    \subfloat[ECG200 dataset]{
        \includegraphics[width=0.66\columnwidth]{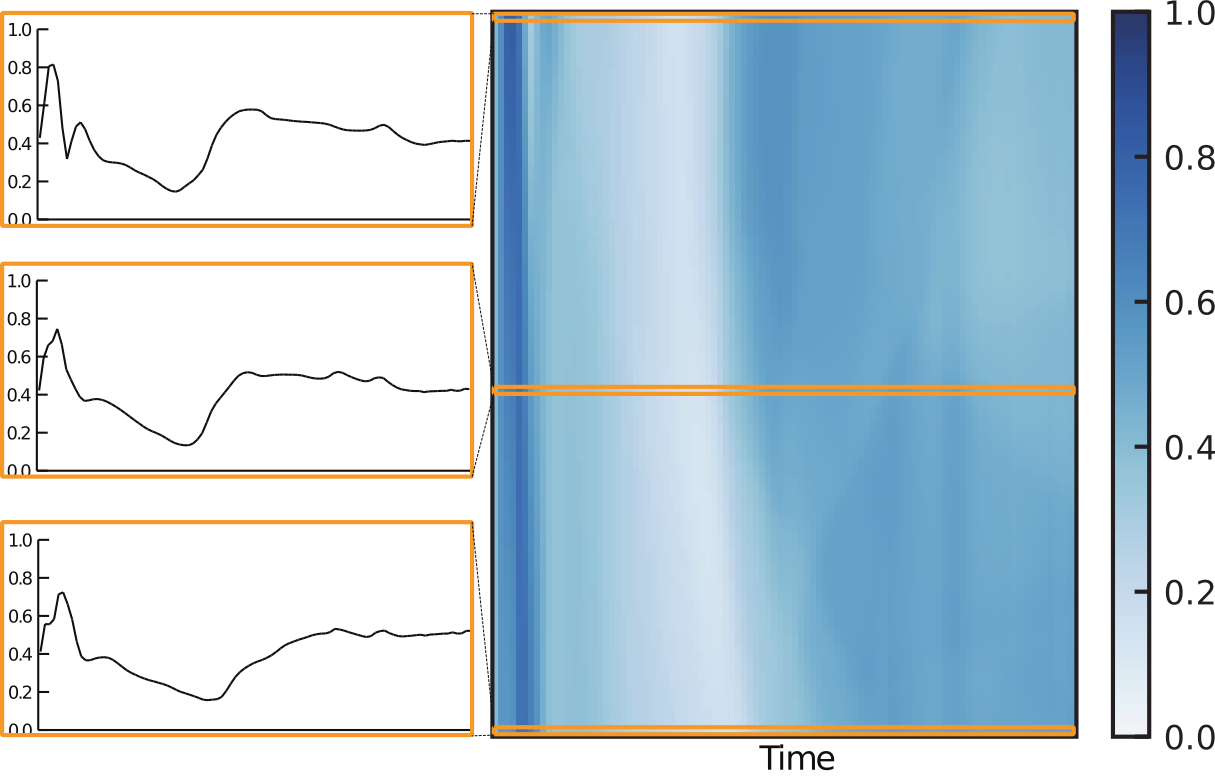}}
    \subfloat[TwoLeadECG dataset]{
        \includegraphics[width=.66\columnwidth]{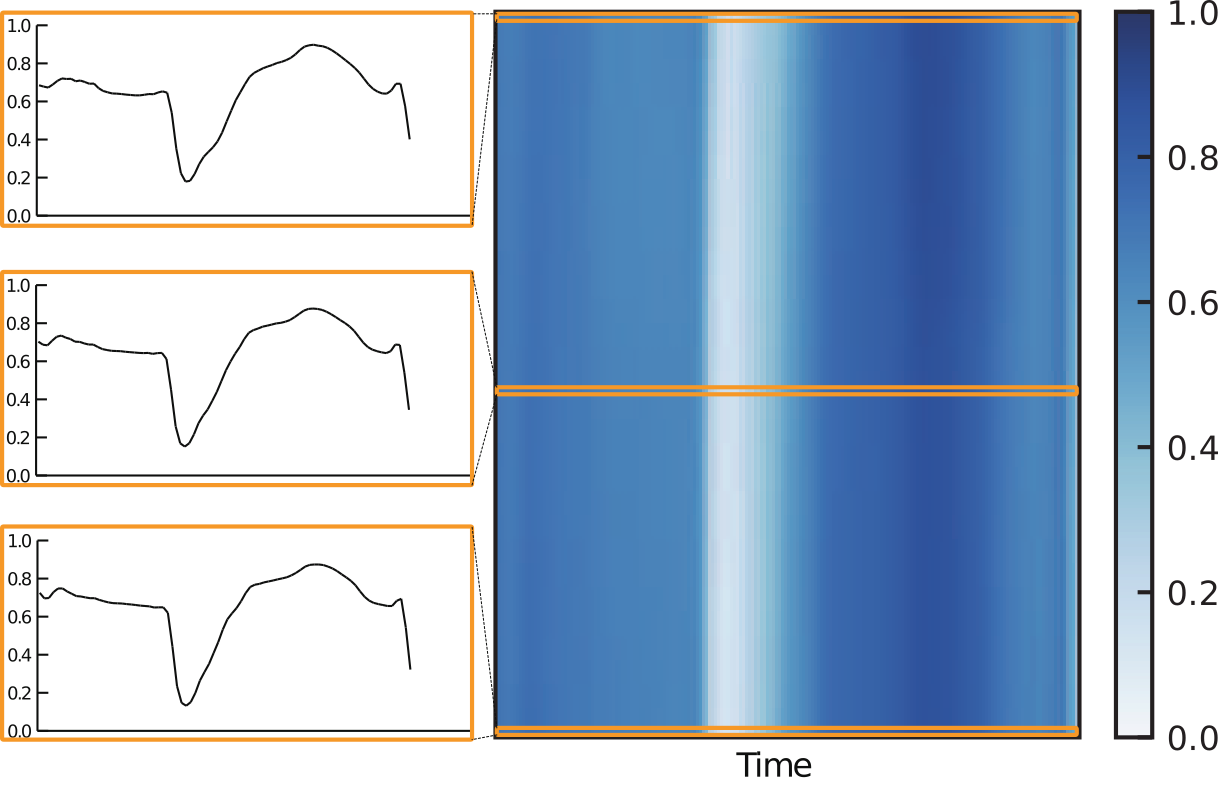}}
    \subfloat[EEG dataset]{
        \includegraphics[width=.66\columnwidth]{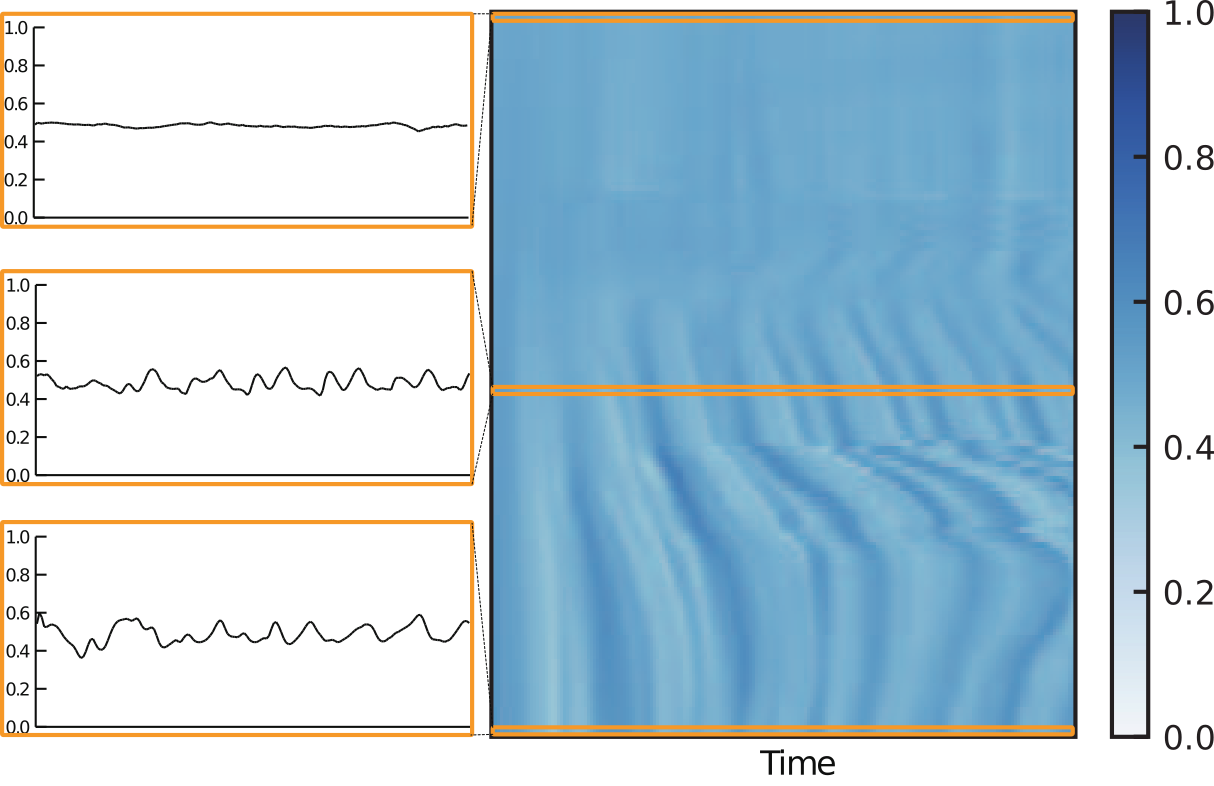}}
    \caption{Results of the interpolation of class labels on the ECG200, TwoLeadECG, and EEG datasets. The top and bottom columns are data generated from fixed random sequence $z$ and class labels $\bm{c}^{(1)}$ and $\bm{c}^{(100)}$, respectively. The vertical axis follows the change in class label $c$, and the horizontal axis shows the data point of each time-series.}
    \label{fig:walk}
\end{figure*}

Fig. \ref{fig:walk} shows the result of the interpolation of class labels by the proposed method.
In these figures, the amplitude of each generated data is shown as a heat map, and the vertical axis shows class label $\bm{c}$.
Figs. \ref{fig:walk}(a), (b), and (c) are the results obtained using the proposed method trained by the ECG200, TwoLeadECG, and EEG datasets, respectively.
In the heat map in the middle of each figure, the results of $G(\bm{z}, \bm{c}^{(1)}), G(\bm{z}, \bm{c}^{(2)}), \cdots , G(\bm{z}, \bm{c}^{(100)})$ in order from the top are shown.

\subsection{Analysis of Latent Variable Space and Control of Generated Data}
The latent variable space was analyzed to control the characteristics of the generated data,.
In the GAN-based method, to generate data with certain characteristics, it is necessary to find input data with the desired characteristics manually from a large number of pairs of generated and input data.
This is because there is no direct parameter for controlling the characteristics of the generated data.
However, this task is highly time consuming and undesirable.
Therefore, it is better to automatically control the behavior of the GAN-based method.
The behavior of the proposed method can be understood by analyzing the input--output relationship.

Canonical correlation analysis (CCA) was conducted to analyze the latent variable space.
CCA is a method of analyzing the interrelationship between two variable groups. 
It linearly converts each variable group into a variable group with the maximum correlation.
CCA determines the transformation, $\bm{w}_x, \bm{w}_y$, and it is defined as
\begin{align}
\left( \bm{w}_{x}, \bm{w}_{y} \right) = \argmax_{\bm{w}_x, \bm{w}_y} \rho \left( \bm{X}, \bm{Y}, \bm{w}_{x}, \bm{w}_{y} \right), \\
\rho \left( \bm{X}, \bm{Y}, \bm{w}_{x}, \bm{w}_{y} \right) = \frac{\bm{X}^T \bm{w}_x \cdot \bm{Y}^T \bm{w}_y}{ \|\bm{X}^T \bm{w}_x \| \|\bm{Y}^T \bm{w}_y \| }. \notag
\end{align}
Even though there is a limit in linear CCA, it was performed for obtaining a broad estimate of the behavior of the proposed method.

In this experiment, CCA was performed on the pairs of the input data and the variables extracted from the generated data.
The various variables extracted from the generated data were the maximum value, the point of the maximum value, the minimum value, the point of the minimum value, maximum-to-minimum interval length, mean amplitude, and the mean frequency on the ECG200 and TwoLeadECG datasets.
In case of the EEG dataset, mean amplitude, standard deviation, median, and mean frequency were extracted from the generated data.
Then, the canonical loadings were obtained, which indicate the contribution rate of the original variable groups to the converted variable groups.

The generated data were controlled by changing the input data based on the canonical loadings obtained from CCA.
The canonical loadings of the highest canonical correlation coefficient multiplied by a constant ranging from $0$ to $2$ were given to the proposed model as input data, and the characteristics of the generated data were observed.

\begin{figure*}[tbp]
    \centering
      \subfloat[Class 1 of the ECG200 dataset]{
        \includegraphics[width=.95\columnwidth]{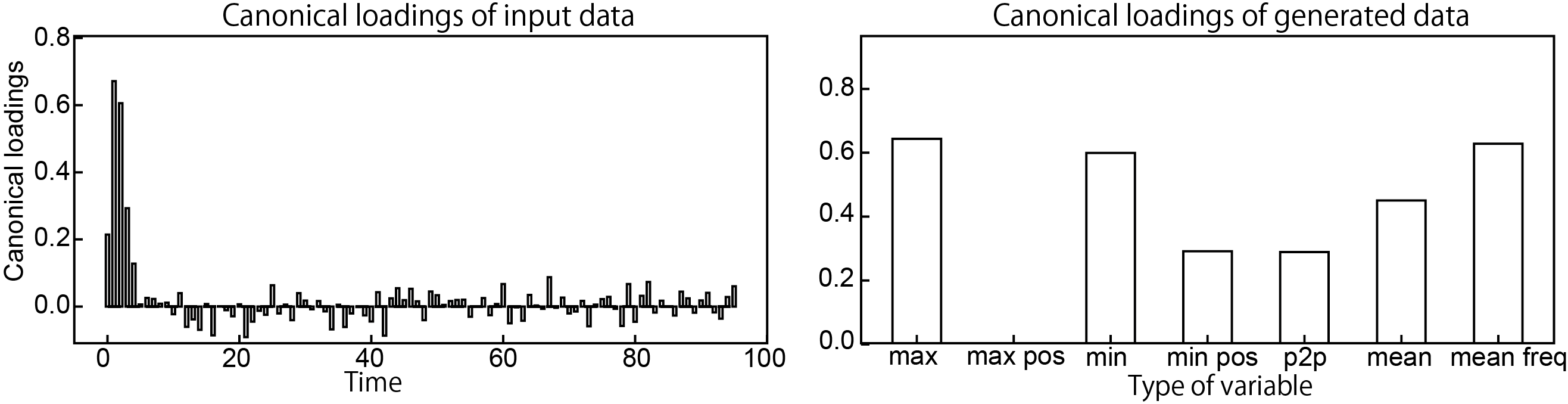}}
        \hspace{.5cm}
    \subfloat[Class 2 of the ECG200 dataset]{
        \includegraphics[width=.95\columnwidth]{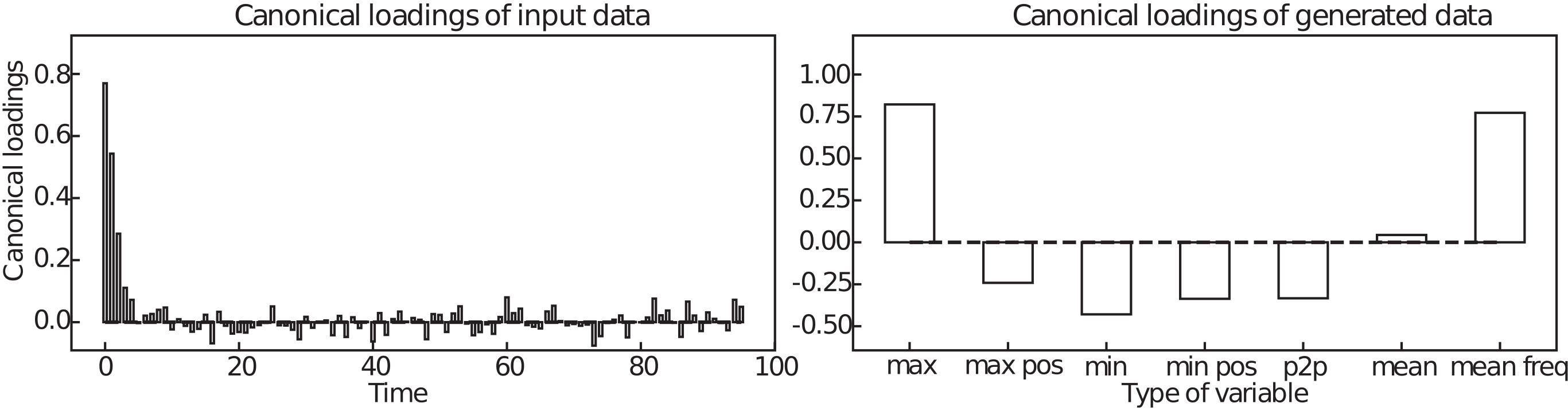}}\\
    \subfloat[Class 1 of the TwoLeadECG dataset]{
        \includegraphics[width=.95\columnwidth]{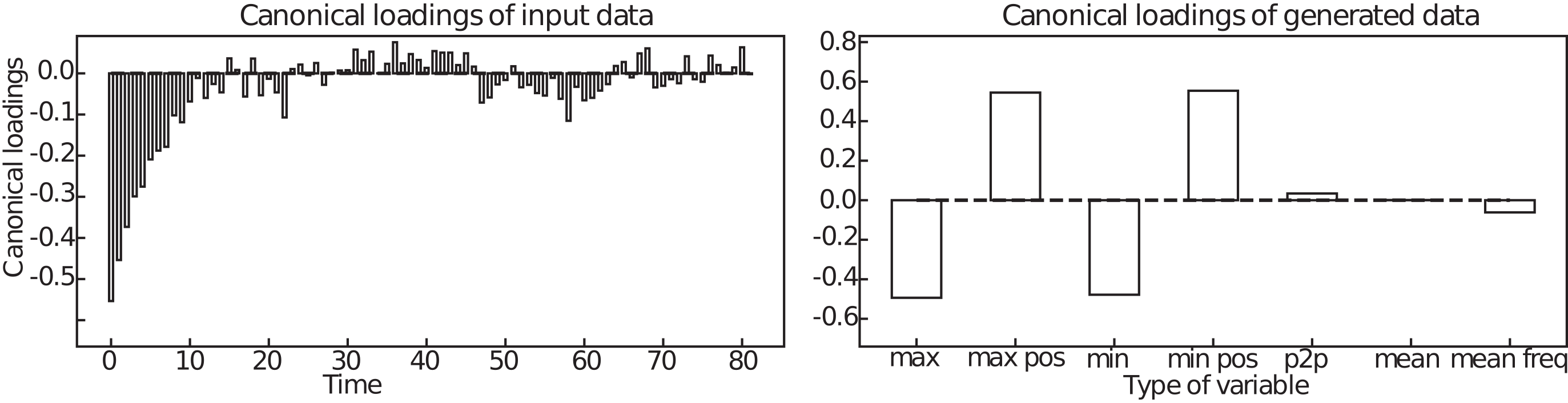}}
        \hspace{.5cm}
    \subfloat[Class 2 of the TwoLeadECG dataset]{
        \includegraphics[width=.95\columnwidth]{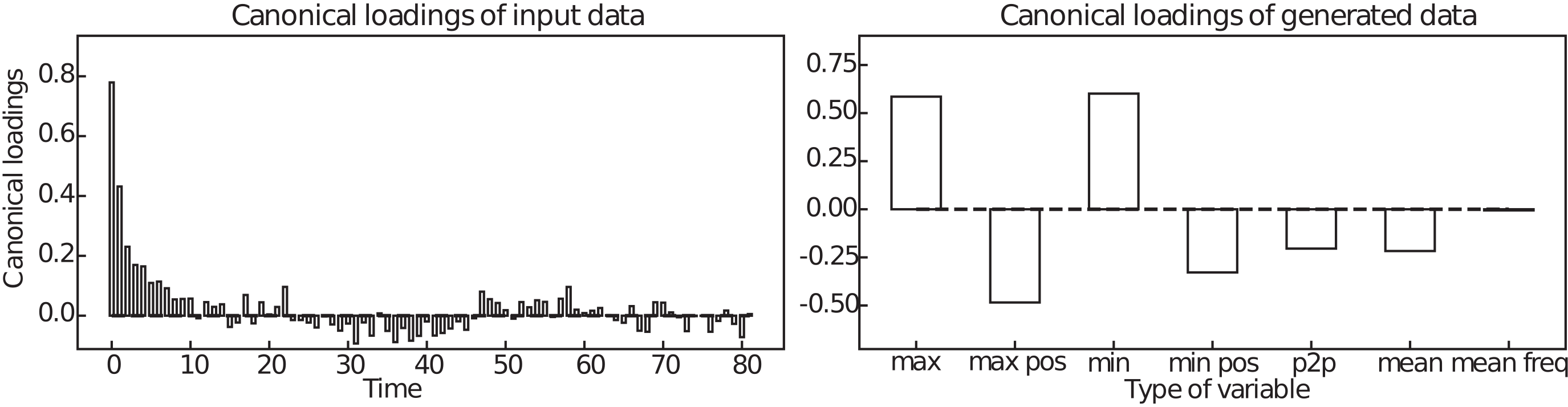}}\\
    \subfloat[Class 1 of the EEG dataset]{
        \includegraphics[width=.95\columnwidth]{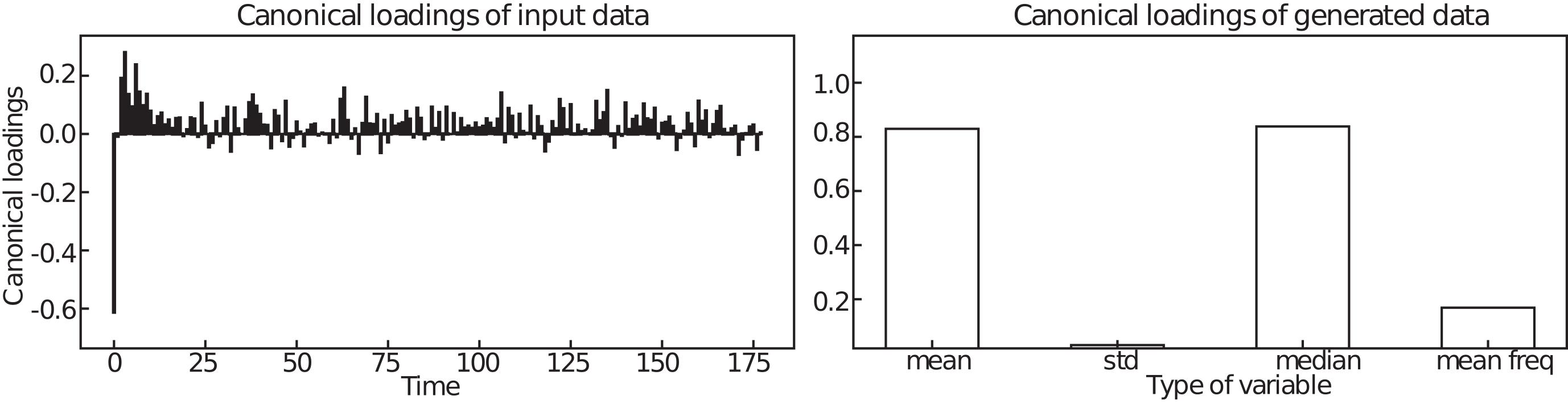}}
        \hspace{.5cm}
    \subfloat[Class 2 of the EEG dataset]{
        \includegraphics[width=.95\columnwidth]{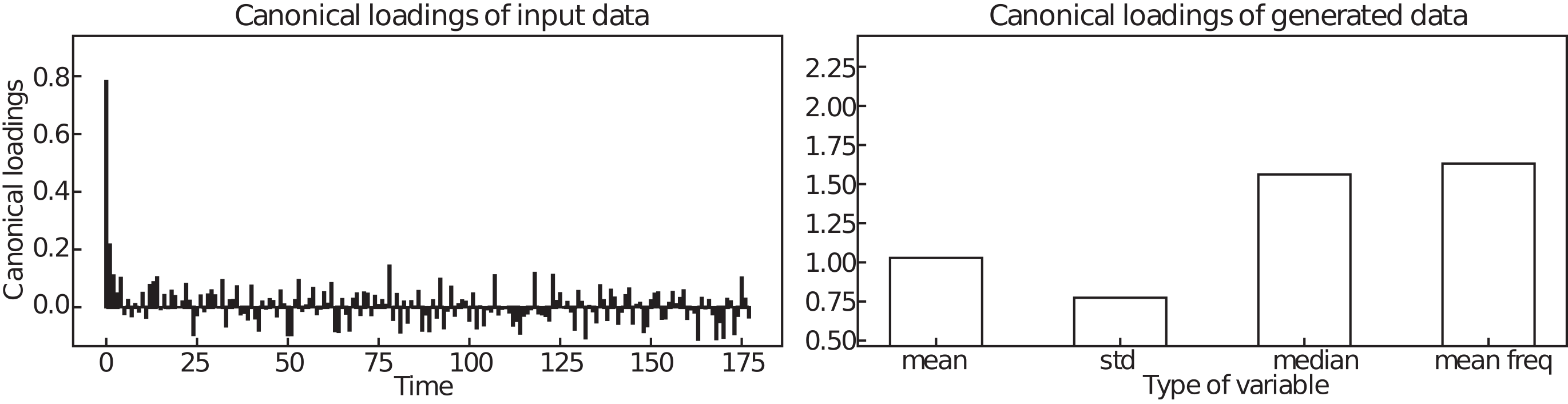}}
    \caption{Results of the input--output analysis of the proposed method obtained from the CCA between the generated data and corresponding input data.}
    \label{fig:c_load}
\end{figure*}

Fig. \ref{fig:c_load} shows the results.
The panels of Fig. \ref{fig:c_load} show the canonical loadings of the ECG200, and TwoLeadECG, and EEG datasets computed by CCA.
The graph is the canonical loading corresponding to the $1$st canonical correlation coefficient.
In each figure, the left side is the canonical loading of the input data and the right side is the canonical loading of the data converted from the generated data.

\begin{figure*}[tbp]
    \centering
      \subfloat[Class 1 of ECG200 dataset]{
        \includegraphics[width=.95\columnwidth]{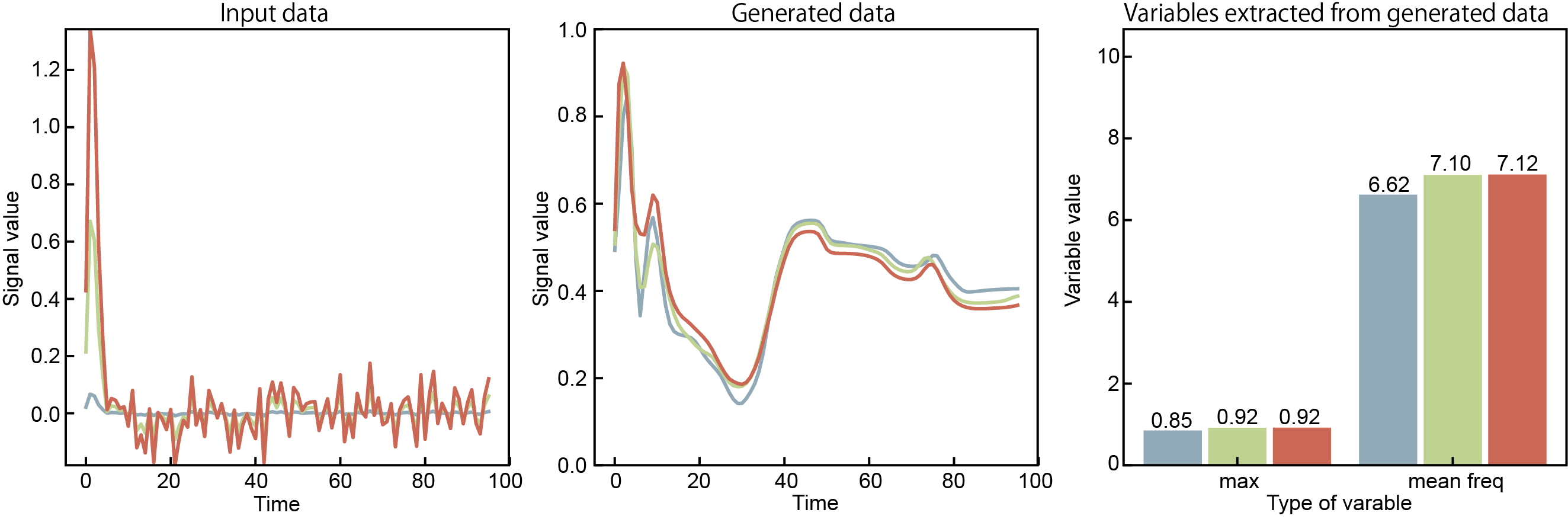}}
        \hspace{.5cm}
    \subfloat[Class 2 of ECG200 dataset]{
        \includegraphics[width=.95\columnwidth]{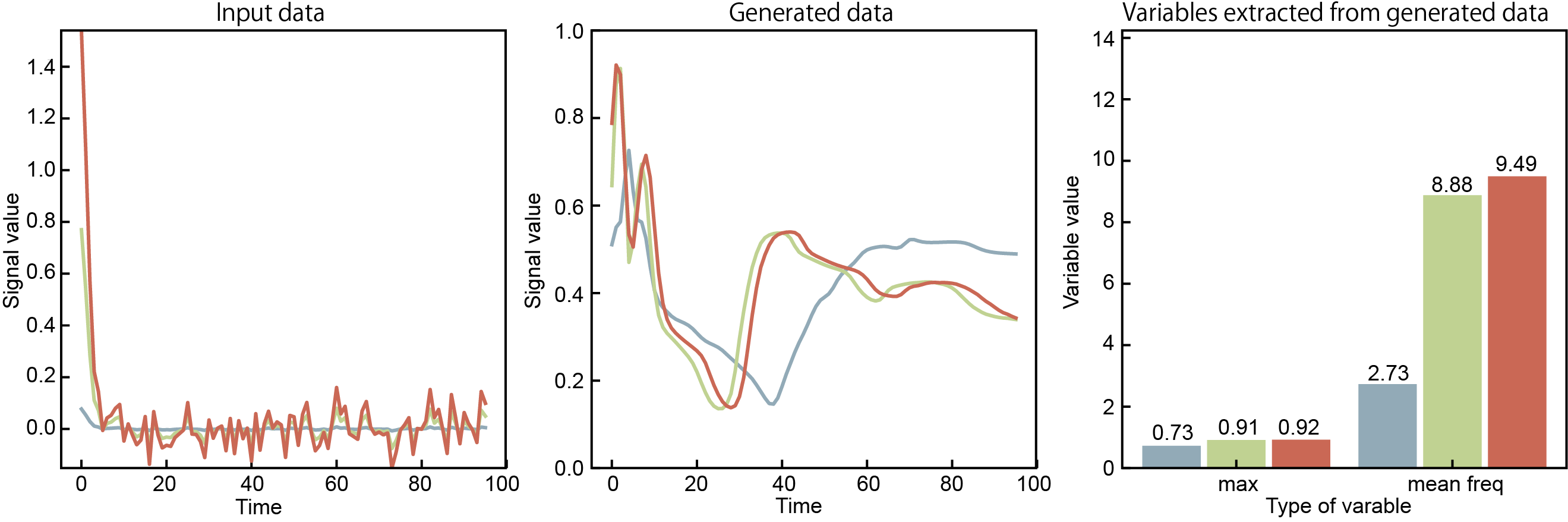}}\\
    \subfloat[Class 1 of TwoLeadECG dataset]{
        \includegraphics[width=.95\columnwidth]{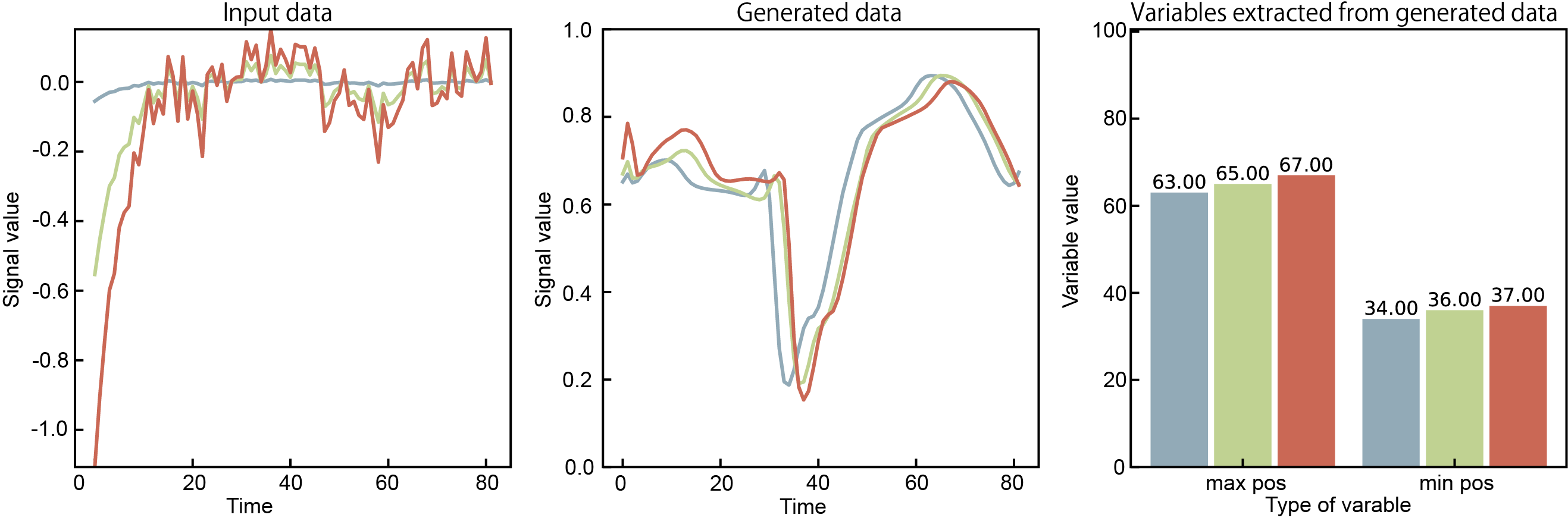}}
        \hspace{.5cm}
    \subfloat[Class 2 of TwoLeadECG dataset]{
        \includegraphics[width=.95\columnwidth]{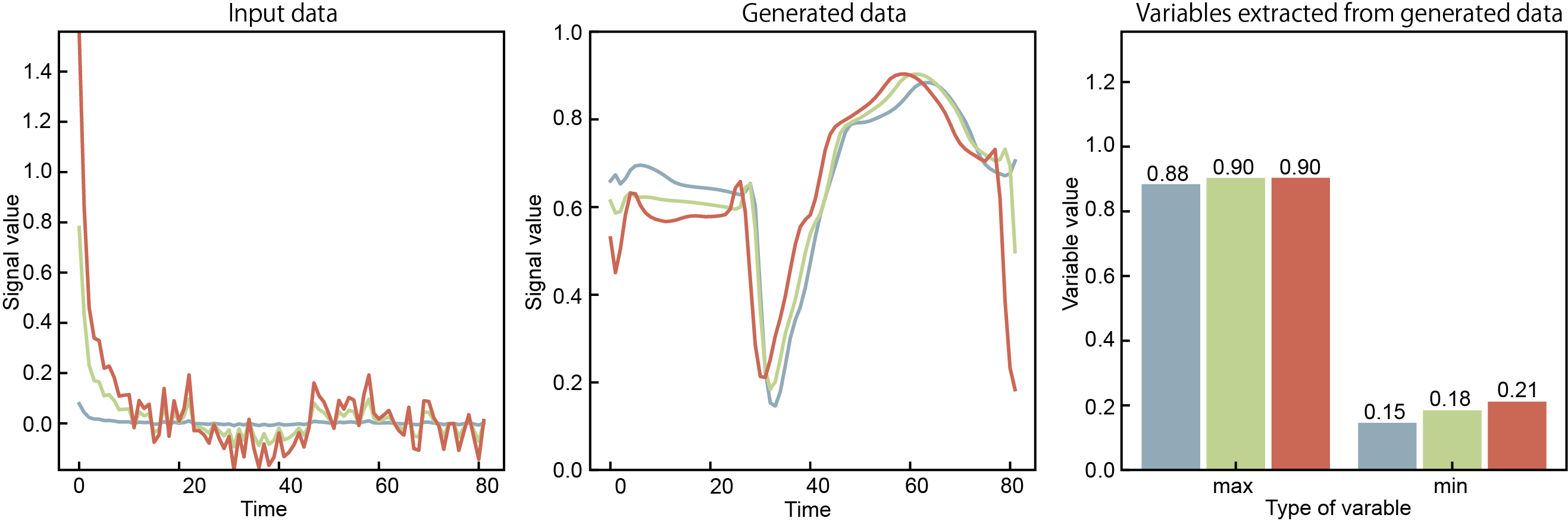}}\\
    \subfloat[Class 1 of EEG dataset]{
        \includegraphics[width=.95\columnwidth]{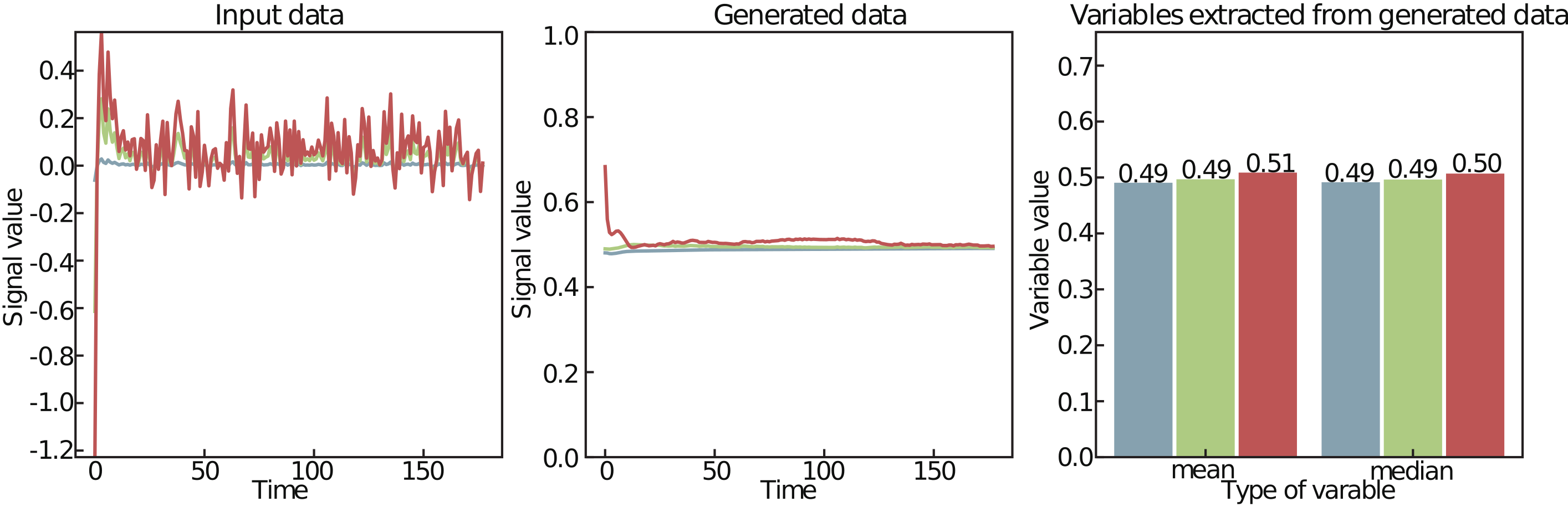}}
        \hspace{.5cm}
    \subfloat[Class 2 of EEG dataset]{
        \includegraphics[width=.95\columnwidth]{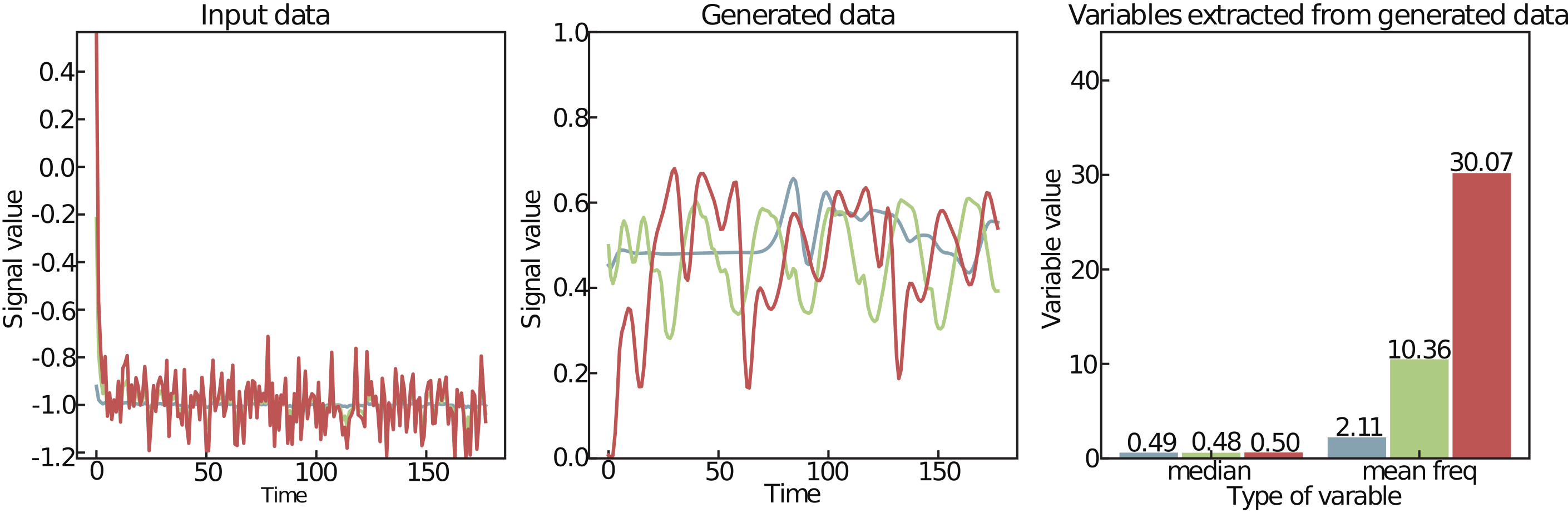}}
    \caption{Example of the result of controlling the generated data based on the CCA results. These input data for the control data were obtained from the first canonical loadings of the input data.}
    \label{fig:control}
\end{figure*}

Fig. \ref{fig:control} shows the results of the attempt to control the generated data based on the first canonical loadings of the input data.
The panels of Fig. \ref{fig:control} show the control results of the ECG200, TwoLeadECG, and EEG datasets.
In each figure, the left, middle, and right parts are the input data based on the canonical loadings, the generated data, and the data converted by the generated data, respectively.

\section{Discussion}
Fig. \ref{gene_data} confirms that the proposed method generates time-series data that have characteristics similar to the original data.
For the ECG200 dataset, the peak close to the initial time point and the rapid decrease and increase around the $30$th time point are retained in the generated data.
For the other datasets, the characteristics of the training data are mostly reproduced.
In addition, from the comparison of classes $1$ and $2$ of the generated data, it is confirmed that the feature is captured for each class.

Fig. \ref{evaluate} quantitatively shows that the quality of the generated data is high because the average similarity between the data generated by the proposed method and the original data is close to that of the original data.
The results of the proposed method are not inferior compared to methods other than our previous method and the HMM.
As the generated data of these methods are obtained by converting the training data in a simple manner, the fact that the result obtained by the proposed method is not inferior compared to the result of these methods further indicates that the quality of the data generated by the proposed method is high.
Furthermore, the results of the proposed method are not inferior compared with our previous method, which trains each class independently.
This result demonstrates that one model can replace the multiple models of our previous method, which should reduce calculation costs.
The number of parameters of the proposed method is $1/C$ times the number of parameters of the conventional method when the number of layers and the number of LSTM units are equal in both methods, where $C$ is the number of classes.

Fig. \ref{fig:walk} confirms that the feature of the generated data can be controlled by the auxiliary information  given at the time of training by the proposed method.
For the ECG200 dataset, the change in amplitude around the initial time point gradually becomes more moderate and the fluctuation of the amplitude around the intermediate time points gradually decreases when the class label changes from class $1$ to class $2$.
For the EEG dataset, the frequency of the generated data increases according to the change in the class label in Fig. \ref{fig:walk}(c).
The transitions of the generated data are reasonable based on the details of each dataset.
From these results, it is confirmed that it is possible to control the characteristics of generated biosignals by training a model using prior information such as class labels.

In Fig. \ref{fig:c_load}, the relationship between the input and generated data is confirmed as the canonical loadings, and the behavior of the model generated by the proposed method can be grasped from the result.
Furthermore, Fig. \ref{fig:control} reveals the effectiveness of the method of controlling the generated data based on the CCA result.
In Figs. \ref{fig:c_load}(a) and (b), there is a strong canonical correlation between the first to fourth time points of the input data and the maximum value and mean frequency of the generated data. 
As a result of control using the canonical loadings shown in Figs. \ref{fig:control}(a) and (b), it is confirmed that the maximum value and mean frequency of the generated data increase according to the change in input data.
In the other datasets, it is confirmed that the characteristics of the generated data are controlled according to the relationship between the input and generated data shown in Fig. \ref{fig:c_load}.
From these results, it is confirmed that the characteristics of the generated data not given as auxiliary information at the time of training, such as the mean frequency and maximum value, can be controlled using the input--output analysis based on CCA.

\section{Conclusion}
In this study, a conditional generation method for time-series data based on GANs was proposed.
In the proposed method, each neural network in a GAN was developed using LSTM units for its hidden layers, thereby allowing for the conditional generation of time-series data according to class labels.
In this method, data similar to training data could be generated without requiring domain-dependent knowledge.

In the experiments, the ability of the proposed method to conditionally generate biosignals was confirmed using three real-world datasets and the controllability of the data generated by the proposed method was verified.
First, the quality of the data generated using each method was quantitatively evaluated using similarity based on the DTW distance.
The results showed the similarity between the original and generated data.
Next, to verify the controllability of the generated data, the input--output relationship of the model generated using the proposed method was analyzed through CCA, and input data changed based on the CCA results were applied to the model.
The input data changed using the CCA results were shown to generate data with the intended changes in the characteristics.
To the best of our knowledge, this study is the first attempt to control the characteristics of the generated data using an analysis of the results of the data generated by a GAN.

A few limitations exist in this study.
First, the learning termination condition of the proposed method cannot be uniquely determined because the loss value output from a GAN does not indicate its learning progress.
Second, hyperparameters must be tuned because the quality of the generated data may vary depending on the hyperparameters.
Third, the characteristics that do not change considerably in the training dataset cannot be controlled using the generated data.
Fourth, it requires a substantially long time to train the model of the proposed method.
Finally, the proposed method may not be able to reproduce high frequency components because the LSTM behaves like a low-pass filter \cite{Bengio2013}.

\end{document}